\documentclass[sigconf, 10pt]{acmart}

\usepackage{array}
\usepackage{booktabs} 
\usepackage{enumerate}
\usepackage{stfloats}
\usepackage{footmisc}

\usepackage{amsmath,amsthm,amssymb,mathtools}
\usepackage{bm}
\usepackage{bbm}

\usepackage[subtle]{savetrees}

\usepackage{caption}
\usepackage{subcaption} 
\usepackage{enumitem}
\usepackage{diagbox}
\usepackage{makecell}
\usepackage{arydshln}
\usepackage{multirow}
\usepackage[noend]{algpseudocode}
\usepackage{algorithm}
\usepackage{color}
\usepackage[export]{adjustbox}
\usepackage{pifont}
\usepackage{bbding}
\usepackage{wasysym}
\usepackage{graphicx}
\usepackage{lipsum} 

\definecolor{gray}{gray}{0.95}
\usepackage{listings}
\usepackage{colortbl}
\usepackage{xcolor}
\usepackage{xspace}

\usepackage[table]{xcolor}

\definecolor{mauve}{rgb}{0.58, 0.44, 0.86}

\newcommand{\workname}{CUHK-X\xspace}

\newcommand{\HAU}{HAU}
\newcommand{\HARn}{HARn}

\theoremstyle{remark}

\usepackage{tikz}
\usepackage{multicol}
\makeatletter
\def\and{\\}
\makeatother

\acmYear{2022}
\acmPrice{15.00}
\acmConference[TBD]{The 24th ACM International Conference on Mobile Systems, Applications, and Services}{TBD}{TBD}

\begin{document}
    \title[\workname]{
    A Large-Scale Multimodal Dataset and Benchmarks
    for Human Activity Scene Understanding and Reasoning}

    \author{Siyang Jiang$^{1}$, Mu Yuan$^{1}$, Xiang Ji$^{1}$, Bufang Yang$^{1}$}
    \author{Zeyu Liu$^{2}$, Lilin Xu$^{3}$, Yang Li$^{1}$, Yuting He$^{1}$, Liran Dong$^{1}$, Wenrui Lu$^{1}$}
    \author{
    Zhenyu Yan$^{1}$, Xiaofan Jiang$^{3}$, Wei Gao$^{4}$, Hongkai Chen$^{1,}$\textsuperscript{\Envelope}, Guoliang Xing$^{1,}$\textsuperscript{\Envelope}
    }
          
    \affiliation{%
    \institution{$^{1}$The Chinese University of Hong Kong, Hong Kong, 
    $^{2}$University of Illinois Urbana-Champaign, United States,} 
    \institution{$^{3}$Columbia University, United States, 
    $^{4}$University of Pittsburgh, United States.
    }
    }
\renewcommand{\shortauthors}{S. Jiang et al.}


\begin{abstract} 
Multimodal human action recognition (HAR) utilizes complementary data for activity classification. 
Built on traditional HAR tasks, recent advances in 
Large Language Models (LLMs) enable detailed descriptions and causal reasoning of human actions, advancing new tasks of human action understanding (HAU) and human action reasoning (HARn). 
However, most LLMs, especially multimodal Large Vision-Language Models (LVLMs), struggle with modalities other than RGB images, like depth, IMU, or mmWave, due to a lack of large-scale \texttt{<data,caption>} datasets in these task domains. Existing HAR datasets provide only coarse-grained \texttt{<data,label>} annotations, insufficient for depicting the detailed action dynamics required in HAU and HAR tasks.\footnote{We focus on two categories of \texttt{<data,ground\;truth>} data pairs: (1) \texttt{<data,label>}, where the \texttt{label} is a discrete category, and (2) \texttt{<data,caption>}, where the \texttt{caption} is a textual description.}
Simply combining annotations and generating captions with LLMs often lacks the necessary logical and spatiotemporal consistency. 

In this paper, we introduce \workname, a large-scale multimodal dataset and benchmarks for HAR, HAU, and HARn.  It includes 58,445 samples of 40 actions performed by 30 participants across two indoor environments, covering diverse daily scenarios. To
address the challenge of spatiotemporal inconsistencies in
captions, we propose a prompt-based scene creation method
that leverages LLMs to generate logically connected activity
sequences. \workname also includes three benchmarks with six tasks to evaluate state-of-the-art models. 
Experimental results show average accuracies of 76.52\% for HAR, 40.76\% for HAU, and 70.25\% for HARn.
This large-scale multimodal dataset aims to empower the research community to apply, develop, and adapt data-intensive learning techniques for a wide range of human activity-related tasks. The project page and code is available at {\color{cyan} \url{https://openaiotlab.github.io/CUHK-X/}} and {\color{cyan} \url{https://github.com/openaiotlab/CUHK-X}}, respectively.
    \end{abstract}

    \settopmatter{printacmref=false}
    \renewcommand{\footnotetextcopyrightpermission}[1]{}

     \maketitle

    \vspace{-0.1in}
\section{Introduction} 
\label{sec:intro}

\begin{figure}[t]
    \centering
    \includegraphics[width=.98\linewidth]{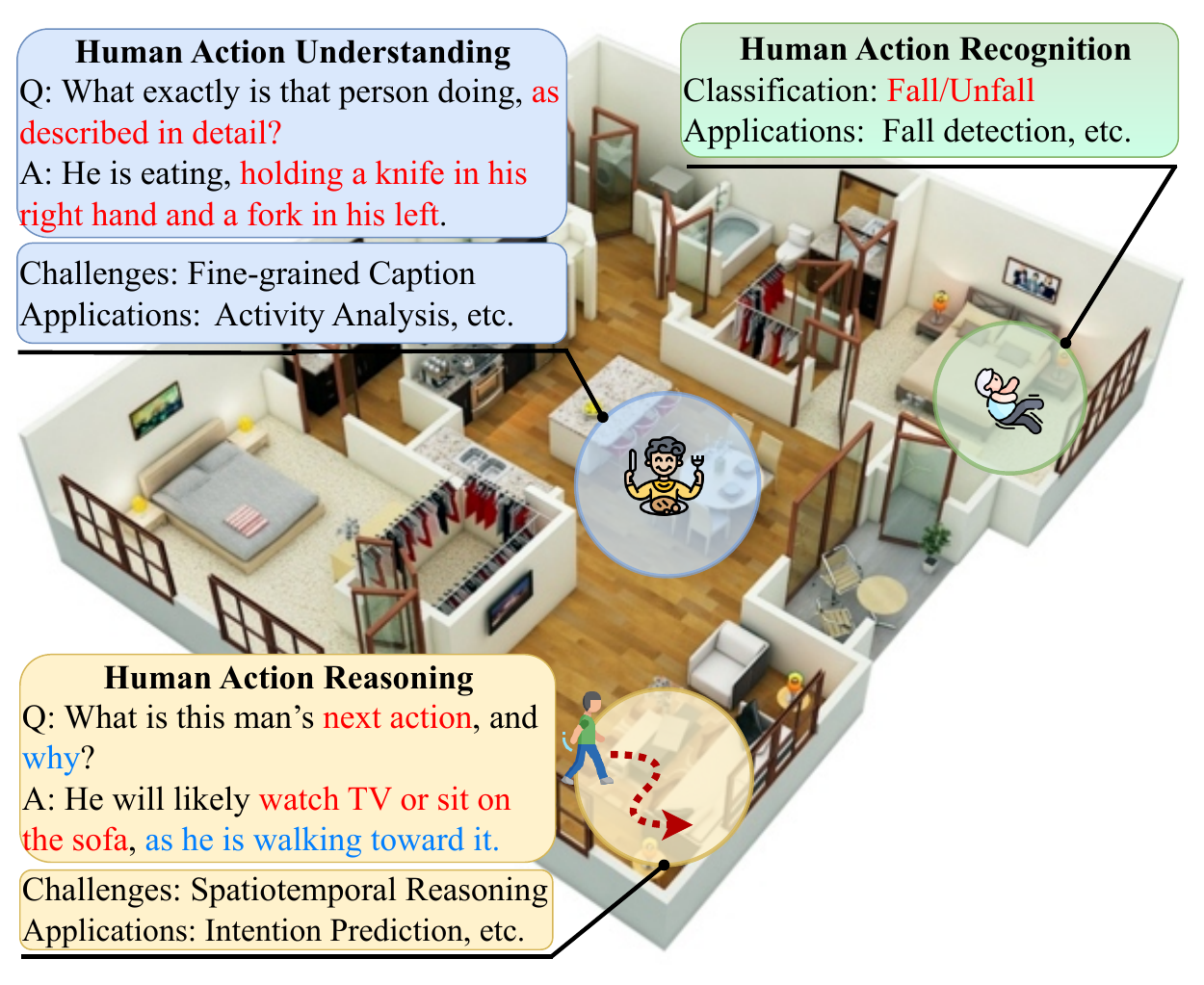}
    \vspace{-10pt}
    \caption{\workname captures a multi-room home environment and supports three tasks: HAR (classification), HAU (captioning), and HARn (intention prediction).
    It integrates diverse modalities, including RGB, depth, thermal, infrared, IMU, skeleton, and mmWave.
 }
    \vspace{-10pt}
    \label{fig:cuhk-x}
\end{figure}
In recent years, human action recognition (HAR) tasks have advanced significantly, leveraging artificial intelligence to classify human activities from multimodal sensory data~\cite{ouyang2022cosmo,zhang2025demo}.
Beyond classification, Human Action Understanding (HAU) and Human Action Reasoning (HARn) tasks provide richer and more detailed descriptions of human activities, enabling diverse applications across various domains such as healthcare, daily living assistance, and surveillance~\cite{yang2025socialmind,wu2024demo,jiang2024artfl}. 
For example, in the management of Alzheimer's disease, a coherent understanding of a patient's longitudinal behaviors is crucial for monitoring daily routines, providing timely caregiver support, and preventing accidents~\cite{qi2025alzheimer}. 
As shown in Fig.~\ref{fig:cuhk-x}, a traditional HAR task is limited to recognizing isolated human actions, such as ``\emph{sleep}'' or ``\emph{fall}'', and lacks the ability to interpret a continuous sequence of actions. 
In contrast, the HAU task addresses this limitation by understanding and providing natural language descriptions of the sequence of actions.
For example, describing a scene as ``\emph{the subject is eating, holding a knife in his right hand and a fork in his left}'' provides valuable context for the early detection of cognitive decline.
Furthermore, the HARn task infers intentions from sequences of human actions and predicts future actions. 
A typical example is observing ``\emph{a subject is walking toward to sofa}''; subsequent action might be predicted as ``\emph{attempting to watch television  or sit down}'', thereby triggering a preventative intervention.

In practice, human action understanding and intention prediction can rarely done by a straightforward autoregressive process with conventional deep neural networks (DNNs). 
Instead, it requires capability for knowledge representation and logical reasoning that integrate environmental contexts and scene knowledge~\cite{yang2024drhouse}. 
To obtain these capabilities, existing techniques usually fine-tune Large Language Models (LLMs) using high-quality datasets annotated as
\texttt{<data,caption>} pairs.
Furthermore, logical reasoning in LLMs can be explicitly elicited through methodologies such as Chain-of-Thought~\cite{mondal2024kam}, Tree-of-Thought~\cite{yao2023tree}, or Graph-of-Thought~\cite{shin2025enhancing}.

However, most existing HAR datasets provide only coarse-grained \texttt{<data,label>} pairs for RGB images~\cite{wang2024xrf55}, thus unsuitable for fine-tuning LLMs for HAU and HARn tasks. 
Although some recent datasets offer fine-grained \texttt{<data,caption>} pairs for RGB images~\cite{caba2015activitynet,grauman2022ego4d,grauman2024ego}, the fixed fields of views (FOVs) and limited mobility of RGB cameras hinder the timely capture of human behavior in many practical scenarios. 
Moreover, in privacy-sensitive scenarios such as daily home monitoring, RGB images pose risks by potentially containing sensitive personal data.
Consequently, alternative sensor modalities, such as depth, thermal, IMU, and mmWave, are preferable.

Most of current LLMs, particularly Large Vision Language Models (LVLMs), perform robustly with RGB and textual data, but they encounter significant difficulties when applied to other prevalent non-RGB modalities. The primary reason is the notable scarcity of  large-scale datasets with \texttt{<data,caption>} pairs in non-RGB modalities~\cite{jiang2025llm}, as most existing ones are confined to coarse-grained \texttt{<data,label>} annotations~\cite{liu2019ntu,shahroudy2016ntu,zhang2012usc}. A naive approach to obtaining captioned datasets across multiple sensory modalities involves merging unimodal datasets, combining their coarse-grained labels, and using an LLM to generate captions.
However, this method frequently yields captions that lack the essential spatiotemporal consistency. 
For example, directly combining actions like brushing teeth and eating into a single scene is illogical, as these two actions typically occur independently in distinct contexts (e.g., a bathroom versus a dining room). Additionally, when generating captions, LLMs often fail to accurately infer humans' behavioral contexts from the given coarse-grained labels, due to their limited representational capacity. 
This often results in incomplete, inaccurate, or even misleading captions~(see \S\ref{sec:appendix:motivation} for details).

To address these gaps, we present \workname, a large‑scale multimodal dataset with seven synchronized modalities (RGB, depth, infrared (IR), thermal, skeleton, IMU, and mmWave) and three benchmarks 
for HAU and HARn tasks,
while also supporting conventional HAR tasks. 
The \workname dataset employs a ground-truth‑first (GT‑first) data collection scheme, where target states are predefined before data recording, to ensure the acquisition of precisely aligned multimodal signals.  
To prevent spatiotemporal inconsistencies and guarantee accurate ground truth, the construction of \workname begins with a Scene-based Caption Generation Framework. 
This framework categorizes human actions into seven thematic groups based on the American Time Use Survey (ATUS)~\cite{atus2013,yu2019activitynet}, from which 40 representative actions are carefully selected based on their frequency and relevance in prior benchmark datasets (e.g., HHAR~\cite{stisen2015smart}, UCI~\cite{reyes2016transition}, and Cosmo~\cite{ouyang2022cosmo}).
Subsequently, LLMs are used to logically connect these actions into semantically coherent captions that depict predefined scenes of daily living, such as living rooms, kitchens, bedrooms, and bathrooms.
These captions incorporate varied contexts (i.e., performing actions in a relaxed, calm or hurried manner) to further enrich the narrative coherence. 
Lastly, we employ a human-checking stage to ensure that the generated captions are consistent with the ground truth, by validating physical plausibility and temporal logic. (see overview in \S\ref{sec:appendix:overview-cuhk}.)

Using the generated captions as the ground truth, \workname comprises over 58,445 daily activity samples from 30 participants, captured across two indoor settings using seven modalities, including RGB, depth, thermal, infrared, skeleton, IMU, and mmWave sensors. The sensor suite includes a Goermicro Vzense NYX 650 (depth), Texas Instruments IWR6843ISK (mmWave radar), Hikvision TB4117 (thermal), and five WitMotion WT 9011DCL-BT50 IMUs.  Participants are instructed to understand and act out the generated captions, enabling the collection of high-quality, well-aligned data pairs.

To verify the dataset's utility, we provide benchmarks for six tasks spanning HAR, HAU, and HARn, by evaluating state-of-the-art baselines of DNNs and LLMs to these benchmark tasks. 
These tasks include {(1)} HAR; {(2-5)} HAU tasks (caption comparison, context analysis, sequential action reordering, and action selection); and {(6)} HARn. The HAR benchmark validates the dataset's sufficient knowledge for recognition tasks.
The HAU benchmarks assess caption comparison (against ground truth), context analysis (e.g., inferring speed of action), temporal ordering (for shuffled actions), and action identification (from a predefined set). The HARn benchmark evaluates an LLM's ability to infer intentions, causal relationships, and logical action progression.
For HAR evaluation, we used state-of-the-art recognition models for each sensor modality, and analyze model performance under long-tail distributions and cross subject situations.
For HAU and HARn evaluation, we employed state-of-the-art baselines, including four captioning models (InternVL2.5-2B~\cite{chen2024internvl}, InternVL2.5-8B~\cite{chen2024internvl}, QwenVL2.5-3B~\cite{Qwen2.5-VL}, QwenVL2.5-7B~\cite{Qwen2.5-VL}) and two reasoning models (VideoLLaVA-7B~\cite{lin2023video} and VideoChatR1-7B~\cite{li2023videochat}).
The goal of these benchmarks is to explore the tasks performance and differentiability over different models and modalities (see \S\ref{sec:appendix:overview-cuhk} for overview illustration).

Experimental results demonstrate that fine-tuning models on \workname significantly improves HAR accuracy compared to using pre-trained models alone, confirming that the dataset provides the necessary knowledge.
Specifically, we achieve an average accuracy of 76.52\% across seven modalities for HAR. 
Additionally, we achieve an average accuracy of 40.76\% (max 50.52\%) across all HAU tasks.
Moreover, we achieve an average accuracy of 70.25\% (max 90.30\%) across three vision modalities for HARn.  
These results confirm that \workname enables robust benchmarking and bridges the key gaps in existing datasets. The main contributions are summarized as follows:
\begin{itemize}
    \item We introduce \workname, a large-scale multimodal dataset comprising 58,445 samples collected from 30 participants using seven sensor modalities (RGB, depth, thermal, Infrared, IMU, mmWave, and skeleton) across two real-world environments.
    It provides diverse and realistic activity data and captions for advanced research in HAR, HAU, and HARn.
    \item To address challenges in logical consistency and spatiotemporal representation, we propose a prompt-based scene creation that leverages prompt-driven LLMs to generate logically connected actions in daily activity scenes with a human checking stage.
    \item We establish three benchmarks containing six tasks to systematically evaluate state-of-the-art baselines.
   Through rigorous analysis of tasks performance and differentiability over different models and modalities, we position \workname as a cornerstone dataset for advancing research in HAR, HAU, and HARn.
\end{itemize}
 
    \section{Motivation Study}

This section outlines the tasks of HAU and HARn, followed by a discussion on the limitations of existing datasets in supporting these tasks.

\subsection{Applications of \HAU~and \HARn}
\label{sec:motivation:limitation-labels}

The \workname dataset can be applied to HAU and HARn tasks across various domains, including smart health~\cite{yang2024drhouse}, smart home~\cite{chen2022rfcam}, and disease intervention~\cite{ouyang2024admarker}.
It enables continuous, longitudinal monitoring and analysis of user behavior, such as in Alzheimer’s Disease monitoring~\cite{chen2019developing}, Parkinson’s Disease management~\cite{sun2024digital}. Beyond healthcare, \workname can also support smart home systems by enhancing comfort and energy efficiency. 
For example, it can use the predicted user actions to adjust room lighting or thermostat settings, thereby optimizing energy consumption while maintaining a comfortable environment. 
This enriched holistic understanding, facilitated by \workname, is critical not only for improved caregiving but also for creating smarter and more efficient living spaces.

\subsection{Limitations of Existing Datasets}
Existing multimodal datasets often suffer from key limitations, such as small subject pools (e.g., USC~\cite{zhang2012usc}, Shoaib~\cite{shoaib2014fusion}, HHAR~\cite{stisen2015smart}) and a restricted range of activities (e.g., UTD~\cite{chen2015utd}, mRI~\cite{an2022mri}, Thermal-IM~\cite{ThermalIM2023}).  
While datasets like NTU-RGBD~\cite{liu2019ntu} and Ego-Exo4D~\cite{grauman2022ego4d} are more extensive, they frequently lack modality diversity (e.g., most miss IR, thermal, or captions), as summarized in Table~\ref{table:summary-datasets-1}. 

\subsubsection{Limitations of existing coarse-grained HAR datasets}
As shown in Table~\ref{table:summary-datasets-1}, many earlier datasets, such as USC~\cite{zhang2012usc}, Shoaib~\cite{shoaib2014fusion}, and HHAR~\cite{stisen2015smart}, are constrained by small participant numbers (fewer than 15) and a narrow range of activities (e.g., only 6–12 actions). 
Similarly, datasets like Thermal-IM~\cite{ThermalIM2023} and UTD~\cite{chen2015utd} involve too few participants or a limited number of activities. 
Some recent datasets, such as NTU-60/120~\cite{liu2019ntu,shahroudy2016ntu}, and PKU-MMD~\cite{liu2017pku}, include larger participant cohorts (e.g., 66 people in PKU-MMD) and more activity classes (e.g., 60 actions in NTU-60). 
However, they primarily focus on RGB and skeleton data, overlooking essential modalities like thermal, IR, and IMUs.
For instance, action recognition relying solely on RGB data becomes challenging under occlusion or when a person faces away from the camera. 
Thus, a major limitation of coarse-grained datasets is their inability to provide sufficient detail for HAU or HARn tasks.

\subsubsection{Limitations of existing fine-grained HAU datasets} 
\label{sec:moti:fg-HAU-dataset}
\begin{figure}[t]
    \centering
    \begin{subfigure}[b]{.48\textwidth} 
        \centering
        \includegraphics[width=\textwidth]{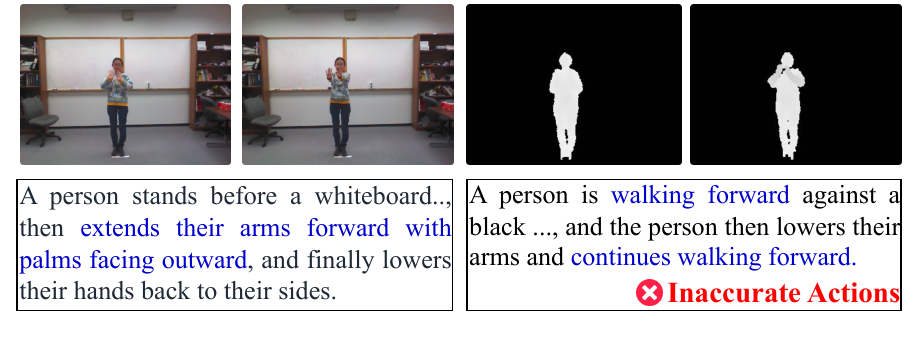} 
        \vspace{-20pt}
        \caption{UTD Dataset}
        \label{fig:motivation:limit_LLMs_rgb}
    \end{subfigure}
    \hfill
    \begin{subfigure}[b]{.48\textwidth} 
        \centering
        \includegraphics[width=\textwidth]{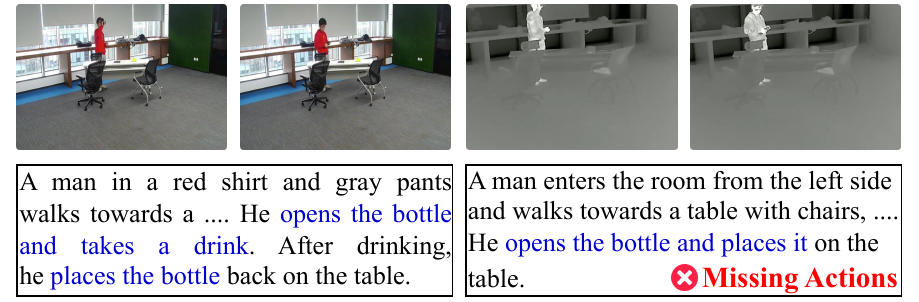} 
        \vspace{-15pt}
        \caption{Thermal-IM Dataset}
        \label{fig:motivation:limit_LLMs_depth}
    \end{subfigure}

    \begin{subfigure}[b]{.48\textwidth} 
        \centering
        \includegraphics[width=\textwidth]{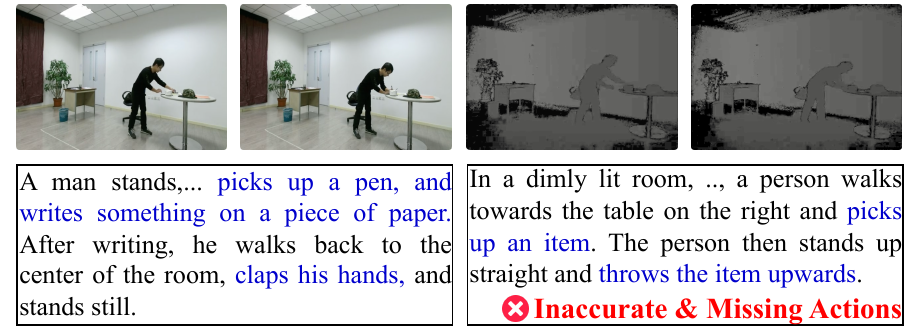} 
        \vspace{-15pt}
        \caption{PKU-MMD Dataset}
        \label{fig:motivation:limit_LLMs_ir}
    \end{subfigure}
    \vspace{-20pt}
    \caption[]{Limitations of SOTA LVLMs in HAU tasks.}
    \label{fig:motivation:limit_LLMs_1}
    \vspace{-15pt}
\end{figure}

Previous fine-grained datasets, such as Ego-4D~\cite{grauman2022ego4d} and Ego-Exo4D~\cite{grauman2024ego}, enhance model capabilities for detailed human action understanding by providing rich data descriptions. Nonetheless, these datasets are limited in modality coverage.
As shown in Table~\ref{table:summary-datasets-1}, Ego-4D and Ego-Exo4D include RGB data but lack other critical modalities, such as depth, thermal, infrared, and skeleton.  Moreover, state-of-the-art captioning models, such as Tarsier~\cite{wang2024tarsier} and Tarsier2~\cite{yuan2025tarsier2}, are designed specifically for RGB data, leading to suboptimal performance in other modalities.
To illustrate, as shown in Fig.~\ref{fig:motivation:limit_LLMs_1}, we conducted experiments on UTD~\cite{chen2015utd}, Thermal-IM~\cite{ThermalIM2023}, and PKU-MMD~\cite{liu2017pku} datasets using Tarsier2 (see \S\ref{sec:appendix:motivation} for details).

\subsection{Summary}

In summary, existing coarse-grained datasets are unsuitable for HAU and HARn tasks due to insufficient descriptive detail, while existing fine-grained datasets fail to comprehensively cover multiple modalities.
To address these gaps, \workname provides a wider variety of modalities (e.g., RGB, depth, thermal, IR, skeleton, IMU, and captions), a more diverse range of activities, and a larger participant cohort.
This comprehensive dataset not only supports HAR tasks but also enables fine-grained HAU tasks that demand a deeper contextual understanding, as well as reasoning-based tasks like HARn, which require understanding sequential actions and predicting future behaviors.
Most importantly, it establishes a foundational resource for developing robust multimodal systems for sophisticated human behavior analysis in real-world scenarios.

    \section{Ground Truth-First Data Collection Strategy}
\label{sec:gt-first-collect}
This section describes our ground truth-first (GT-first) methodology for human action data collection, positioning it as a practically better alternative compared to conventional data-first approaches to human data collection. 

\subsection{Motivation for a GT-First Strategy}
Data collection and annotation typically follow two paradigms: data‑first~\cite{kuznetsova2020open} and ground‑truth‑first, or GT‑first~\cite{liu2019ntu,yang2023mm,caba2015activitynet,an2022mri,ouyang2022cosmo}. \textit{Data-first} strategy, as the most intuitive method of data collection, first gathers large-scale real-world data from human subjects, and then apply annotations to the collected data afterwards. While conceptually reasonable, these strategies present significant limitations in practice. In particular, since data collection is conducted before annotations and hence lacks specific guidelines, the collected data requires intensive postprocessing and filtering to ensure efficient and proper annotation. These extra efforts make data collection highly ineffective and not scalable, especially when involving large groups of human subjects. In addition, since annotations have to be applied to collected data, it is expensive to fix any early mistakes in the human action set, raising extra privacy and consent risks when recording people and homes.

To avoid these limitations, most existing large-scale datasets, such as ImageNet~\cite{deng2009imagenet} and NTU-RGBD~\cite{liu2019ntu}, follow the GT-first strategy for data collection. Being different from the data-first strategy, the GT-first strategy defines the label space, annotation rules, and target scenes in advance, and then collects data based on those definitions. Prioritizing ground truth yields clear benefits: 1) \emph{focus}, as only planned cases are recorded; 2) \emph{efficiency}, as annotators verify checklists rather than create labels from scratch; 3) \emph{scalability}, as the same collection scripts and annotation rules can be reused across sites.

However, the approach would introduce bias compared to the data-first approach. In the following, our objective is to analyze and mitigate three inherent biases of GT-first approaches in \workname, i.e., \emph{coverage, diversity, and discrepancy}, through several targeted strategies including: (1) refining class coverage within defined boundaries, (2) enhancing variation via intra- and inter-class combinations coupled with linguistic enrichment, and (3) incorporating human-in-the-loop verification to ensure physical and logical coherence. These methodological components are elaborated in~\S\ref{sec:caption_generation}.

\subsection{Bias in GT-First Data Collection}
\label{sec:motivation:bias-GT}
GT-first approaches, i.e., those that derive supervision or text directly from predefined actions and associated templates, offer clear benefits in controllability and reproducibility to obtain the GT. In \workname, we adopt GT-first approach to obtain the ground truth. 
However, it also induces systematic artifacts that reflect the prior encoded by the ontology and the templating process. In particular, they may bias models toward (i) \textit{Coverage}, (ii) \textit{Diversity} and (iii) \textit{Discrepancy}. In the following, we justify the biases in \workname.

\subsubsection{Coverage} We define coverage as the number of classes included in the dataset. It is hard for any dataset to include all human actions. In practice,  some labels mean more than one thing, and they are not consistent across datasets.  Consequently, in \workname, we adopt a closed-world objective, i.e., we focus on a fine-grained subset of actions curated from prior research and informed by our experimental evidence. In particular, we coarse-grained our action selection using ATUS~\cite{atus2013} and the action frequency across datasets (\S\ref{sec:cg-action-selection}), and then fine-grained the selection of several significant actions based on prior studies (\S\ref{sec:fg-action-selection}). 

\subsubsection{Diversity}  Even when labels are well covered, their realizations can exhibit narrow lexicalization, limited contextual variety, and stereotyped co-occurrence patterns. We therefore define diversity, for each modality, as the extent to which samples within the same class differ from one another. For example, in each modality, instances of the class ``drinking water'' should vary in setting, poses, and previous or co-occurring actions. To mitigate this limitation, we first generate captions from diverse intra- and inter-class action combinations (\S\ref{sec:intra-inter-caption}) and then enrich those captions with linguistic variation to diversify lexical choice, syntax, and contextual framing (\S\ref{sec:enrich-caption}). 

\begin{table}[t]\small
    \centering
    \caption{Evaluation on discrepancy.}
    \vspace{-10pt}
    \begin{tabular}{c ccc}
        \hline
        \textbf{BertScore-F1} &   \textbf{Com-Cap.} &  \textbf{Free-Act.} & \textbf{Instruct-Act.} \\
        \hline
        Task-1    & 93.40\% & 95.50\% & 93.60\%  \\
        Task-2    & 92.90\% & 94.10\% & 93.10\%  \\
        Task-3    & 93.50\% & 94.20\% & 92.70\%  \\
        Task-4    & 92.20\% & 93.80\% & 92.50\%  \\
        \hline
    \end{tabular}
    \vspace{-10pt}
    \label{tab:pre-exp}
\end{table}
\subsubsection{{Discrepancy}.}
In \workname, discrepancy is denoted from two perspectives. Firstly, it is the mismatch between captions generated by an LLM and those authored by humans for the same meaning. To minimize this discrepancy, we incorporate a human checking stage that enforces physical plausibility, logical coherence, and contextual appropriateness (\S\ref{sec:hd-checking-phy}), aiming to bring LLM captions close to human parity. 
Secondly, we consider whether the data reflect key actions performed naturally versus performed by following explicit instructions. Because \workname~focuses on daily activities, we regard any bias introduced by instruction-following as acceptable. In practice, we asked the volunteer to read the instructions first, then perform the action based on their understanding. We also conduct an experiment to assess this effect. 

Specifically, we set a home office setting with four micro-tasks: drinking water (Task-1), listening to music with earphones (Task-2), writing notes (Task-3), and answering a phone call (Task-4). For each task, we recruited three volunteers (10\% of the participants in \workname) to act freely and annotate their recordings. Next, the same volunteers performed the actions following instructions derived from their free actions. We again collected three independent captions.
We compared the semantic similarity of captions using close-source LLMs, i.e., QwenVL3-235B, between the ``free'' and ``instructed'' conditions using BertScore. 
Across all four tasks, the two sets of captions are highly similar, indicating that both conditions effectively capture the underlying action information (Table~\ref{tab:pre-exp}, Com-Cap.). Furthermore, when compared the generated captions against human-annotated ground truth, the results are also accurate, as shown in Table~\ref{tab:pre-exp} (Free-Act. and Instruct-Act.). These findings justify our focus on human actions, and the remaining discrepancies are acceptable within \workname.

\begin{figure}[t]
    \centering
    \includegraphics[width=.95\linewidth]{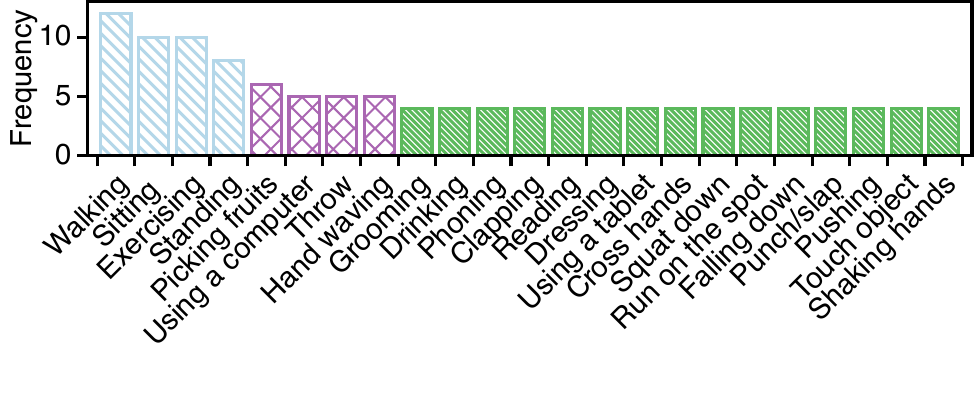}
    \vspace{-30pt}
    \caption{Frequency of the actions among USC~\cite{zhang2012usc}, Shoaib~\cite{shoaib2014fusion}, HHAR~\cite{stisen2015smart}, UTD~\cite{chen2015utd} ActivityNet~\cite{yu2019activitynet}, UCI~\cite{reyes2016transition}, NTU~\cite{liu2019ntu,shahroudy2016ntu}, PKU-MMD~\cite{liu2017pku}, Cosmo~\cite{ouyang2022cosmo}, mRI~\cite{an2022mri}, Thermal-IM~\cite{ThermalIM2023}. }
    \vspace{-10pt}
    \label{fig:action-fre}
\end{figure}

\begin{table*}[!ht]\small
\caption{A summary of the related coarse-grained HAR and fine-grained HAU datasets (\CIRCLE\ indicates inclusion). DEP, THE, IR, SKE denotes depth, thermal, infrared, skeleton modalities, respectively.}
\vspace{-10pt}
\begin{tabular}{c|cccccccccccc}
\toprule
\multirow{2}{*}{\textbf{Dataset}}   & \multirow{2}{*}{\textbf{Years}}     & \multirow{1}{*}{\textbf{\# of}} & \multirow{1}{*}{\textbf{\# of}} & \multirow{1}{*}{\textbf{\# of}}  &  \multirow{2}{*}{\textbf{RGB}} & \multirow{2}{*}{\textbf{DEP}} & \multirow{2}{*}{\textbf{THE}} & \multirow{2}{*}{\textbf{IR}} & \multirow{2}{*}{\textbf{SKE}} & \multirow{2}{*}{\textbf{IMU}} & \multirow{2}{*}{{\textbf{mmWave}}} & \multirow{2}{*}{\textbf{Caption}} \\
&& \textbf{Samples} & \textbf{Subjects} & \textbf{Activities} &&&&&&&& \\
\midrule
USC~\cite{zhang2012usc} & 2012 & 840  & 14  & 12  & \Circle & \Circle & \Circle &  \Circle &  \Circle &  \CIRCLE & \Circle & \Circle  \\
Shoaib~\cite{shoaib2014fusion} & 2014  & 70 &  10  & 7 & \Circle & \Circle & \Circle &  \Circle &  \Circle & \CIRCLE & \Circle & \Circle \\
HHAR~\cite{stisen2015smart}  & 2015 & 3,240  & 9  & 6  & \Circle & \Circle & \Circle &  \Circle &  \Circle &  \CIRCLE  & \Circle & \Circle \\
UTD~\cite{chen2015utd} & 2015 & 3,444 &  8  & 27 & \CIRCLE & \CIRCLE & \Circle &  \Circle &  \CIRCLE &  \CIRCLE & \Circle & \Circle \\        
UCI~\cite{reyes2016transition} & 2016 & 180 & 30  & 6 & \Circle & \Circle & \Circle &  \Circle &  \Circle &  \CIRCLE & \Circle & \Circle \\
NTU-60~\cite{shahroudy2016ntu} & 2016 & 56,880 & 40  & 60 & \CIRCLE & \CIRCLE & \Circle &  \CIRCLE &  \CIRCLE &  \Circle & \Circle & \Circle \\
PKU-MMD~\cite{liu2017pku} & 2017 & 20,000 &  66 & 51 & \CIRCLE & \CIRCLE & \Circle &  \CIRCLE &  \CIRCLE &  \Circle & \Circle & \Circle \\
NTU-120~\cite{liu2019ntu} & 2019 & 114,480 & 106  & 120 & \CIRCLE & \CIRCLE & \Circle &  \CIRCLE &  \CIRCLE &  \Circle & \Circle & \Circle \\
Cosmo~\cite{ouyang2022cosmo} & 2022 & 3,434 & 30  & 14 & \Circle & \CIRCLE & \Circle &  \Circle &  \Circle &  \CIRCLE & \CIRCLE & \Circle \\
mRI~\cite{an2022mri} & 2022 & 300 & 20  & 12   & \CIRCLE & \CIRCLE & \Circle &  \Circle &  \Circle &  \CIRCLE & \CIRCLE & \Circle \\
Thermal-IM~\cite{tang2023happened} & 2023 & 783 & 2  & 24   & \CIRCLE & \Circle & \CIRCLE &  \Circle &  \Circle &  \Circle & \Circle & \Circle \\
MM-Fi~\cite{yang2023mm} & 2023 & 1080 &  40 & 27  & \CIRCLE & \CIRCLE & \Circle &  \Circle &  \CIRCLE &  \Circle & \CIRCLE & \Circle \\
XRF55~\cite{wang2024xrf55} & 2024 & 42,900 &  39 & 55  & \CIRCLE & \CIRCLE & \Circle &  \CIRCLE &  \Circle &  \Circle & \CIRCLE & \Circle \\
\midrule
ActivityNet~\cite{caba2015activitynet} & 2015 & 9,682 & -  & 203  & \CIRCLE & \Circle & \Circle &  \Circle &  \Circle &  \Circle & \Circle & \CIRCLE \\
Ego-4D~\cite{grauman2022ego4d} & 2022 & 5,831 & 923  & 146  & \CIRCLE & \Circle & \Circle &  \Circle &  \Circle &  \CIRCLE &\Circle & \CIRCLE \\
Ego-Exo4D~\cite{grauman2024ego} & 2024 & 5,035 & 740  & 689  & \CIRCLE & \Circle & \Circle &  \Circle &  \Circle &  \CIRCLE & \Circle & \CIRCLE \\
\midrule
\textbf{\workname}  & 2025 & 58,445 & 30  & 40   & \CIRCLE & \CIRCLE & \CIRCLE &  \CIRCLE &  \CIRCLE &  \CIRCLE & \CIRCLE & \CIRCLE \\            
\bottomrule
\end{tabular}
\vspace{-5pt}
\label{table:summary-datasets-1}
\end{table*}

\section{Scene-based Caption Generation}
\label{sec:caption_generation}
\subsection{Prior knowledge-based Action Selection}
\label{sec:cg-fg-action_selection}
\workname~is developed through a labor-intensive collection of multimodal data in real-world scenarios. In this section, we describe how \workname~incorporates typical daily actions with a two-stage selection process to select representative actions. 

\subsubsection{Coarse-grained Action Selections via Predefined Categories and Cross-dataset Frequency}
\label{sec:cg-action-selection}
Firstly, based on the ATUS~\cite{atus2013} activity hierarchy and ActivityNet~\cite{caba2015activitynet}, we categorized the activity classes into seven top-level categories: Personal Care, Eating and Drinking, Household, Caring and Helping, Working, Socializing and Leisure, and Sports and Exercise. \workname~adopts a structured semantic framework that leverages hierarchical relationships between activities, ensuring the selection of typical and comprehensive real-world daily activities. Next, we analyzed the action frequency in 12 popular human action recognition datasets, such as NTU~\cite{shahroudy2016ntu,liu2019ntu}, UTD~\cite{chen2015utd}, and UCI~\cite{reyes2016transition}, which are primarily summarized in Table~\ref{table:summary-datasets-1}. As illustrated in Fig.~\ref{fig:action-fre}, we analyze the high-frequency occurrences within these datasets. Specifically, the total number of action classes is 349, which are consolidated into 127 classes by merging those with similar meanings. However, since most actions appear only once, we focus exclusively on high-frequency actions  (\#frequency > 4)  for further analysis.

\subsubsection{Fine-grained Action Selections via Prior Studies} 
\label{sec:fg-action-selection}

Then, we carefully selected fine-grained, representative actions based on insights from previous research. Specifically, the \textit{Personal Care} category (6 actions) was guided by the prior study~\cite{mlinac2016assessment}. The \textit{Eating and Drinking} (6 actions) and \textit{Household} (5 actions) categories were guided by findings from the previous work~\cite{quinn2011functional}. The \textit{Working} category (6 actions) was inspired by~\cite{anakpo2023impact}, while the \textit{Socializing and Leisure} category (5 actions) was shaped by~\cite{bovckus2023wellness}. Finally, the \textit{Sports and Exercises} (9 actions) and \textit{Caring and Helping} (3 actions) categories were supported by insights from Gerber et al.~\cite{gerber2023practicing}.

\begin{figure}[t]
    \centering
    \includegraphics[width=.88\linewidth]{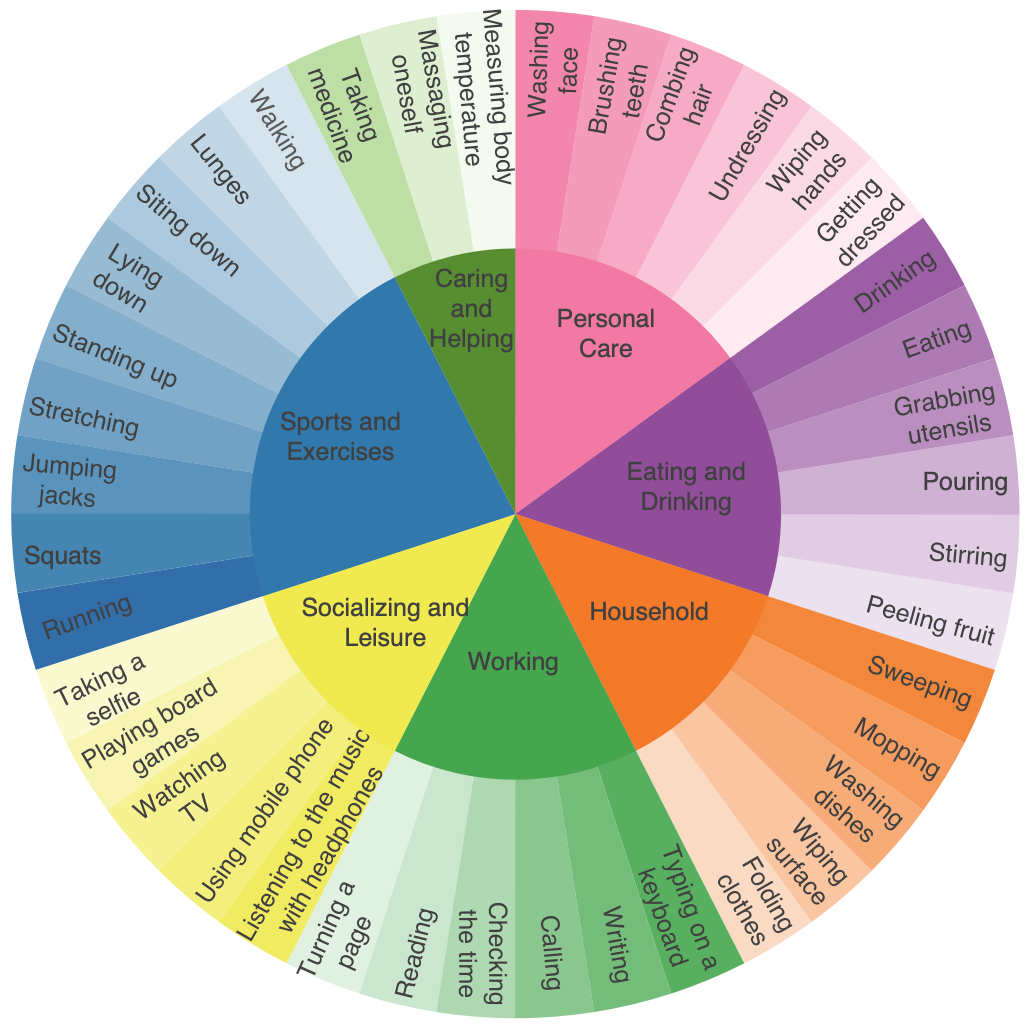}
    \vspace{-10pt}
    \caption{\workname includes 40 actions in 7 categories.}
    \vspace{-15pt}
    \label{fig:overview-actions}
\end{figure}

As shown in Fig.~\ref{fig:overview-actions},  we select 40 actions which are divided into seven categories include the following: \textbf{(1) Personal Care}, which has 6 actions including Washing face, Brushing teeth, Combing hair, Undressing, Wiping hands, and Getting Dressed; \textbf{(2) Eating and Drinking}, which has 6 actions including Drinking, Eating, Grabbing utensils, Pouring, Stirring, and Peeling fruit; \textbf{(3) Household}, with 5 actions including Sweeping, Mopping, Washing dishes, Wiping surface, and Folding clothes; \textbf{(4) Working}, which includes 6 actions including Typing on a keyboard, Writing, Calling, Checking the time, Reading and Turning a page; \textbf{(5) Socializing and Leisure}, with 5 actions including Taking a selfie, Playing board games, Watching TV, Using a phone, and Listening to the music with headphones; \textbf{(6) Sports and Exercises}, which has 9 actions including Walking, Lunges, Siting down, Lying down, Standing up, stretching, Jumping jacks, Squats and Running and \textbf{(7) Caring and Helping}, with 3 actions including Taking medicine, Checking body temperature, and Massaging oneself.

\subsection{Prompt-based Scene Creation with Human-driven Checking}
\label{sec:pb-scene}
In this subsection, we propose a prompt-based scene creation approach designed to logically connect the selected actions (\S \ref{sec:cg-fg-action_selection}) via constructing various daily living scenes.

\subsubsection{Intra- and Inter-categories Caption Generation} 
\label{sec:intra-inter-caption}
Our goal is to connect as many selected actions as possible into a coherent and logical sentence that aligns with everyday life scenarios.
To achieve this, we implement a two-stage prompt design primarily based on the selected actions, ensuring both diversity and relevance in the generated captions. Specifically, we design the prompt to encourage the LLM to combine multiple actions into a cohesive scene within each category. For instance, actions such as ``Washing face,'' ``Brushing teeth,'' ``Combing hair,'' ``Dressing,'' and ``Wiping hands'' can be combined to generate a detailed scene: \emph{The user wakes up, opens the curtains, and stretches (Stretching). The user walks to the bathroom and washes their face (Washing face) with water or facial cleanser, then dries it with a towel. The user picks up a toothbrush, squeezes toothpaste, and begins brushing their teeth (Brushing teeth). After brushing, they rinse their mouth with water and clean the toothbrush. The user uses a comb to carefully brush their hair (Brushing hair), possibly tying it up or styling it. The user quickly wipes their hands (Wiping hands) with a towel or tissue. Finally, the user returns to the bedroom, selects clothes from the wardrobe, and completes the process of getting dressed (Dressing).}  Similarly, actions from other categories are combined via LLMs to create contextually rich captions that reflect realistic and meaningful daily scenarios. This method ensures that the generated captions not only integrate multiple actions logically but also create a natural flow of events that mirrors real-life activity patterns.

\subsubsection{Enriching Captions through Language Diversity} 
\label{sec:enrich-caption}
To further enhance the diversity of captions, we leverage LLMs to enrich them by expanding or substituting their sentence components. Specifically, a sentence is composed of several key elements, including the subject, predicate, object, attribute, adverbial, and complement. In the context of central actions in human action understanding, the subject, predicate, and object are typically predefined. Thus, we use LLMs such as GPT-4o~\cite{hurst2024gpt} or DeepSeek~\cite{liu2024deepseek}, to add more attributes and adverbials. For instance, we can enrich the description of the example in \S\ref{sec:intra-inter-caption} by incorporating adverbs before the verb to provide additional context and nuance. Specifically, instead of a straightforward description, the user can \emph{carefully squeeze a small amount of toothpaste onto the bristles and begin brushing their teeth (Brushing teeth),} with added detail such as ``quickly,'' ``smoothly,'' or ``slowly'' to describe how the action is performed. By enriching the attributes and adverbials, the generated captions provide a more detailed and vivid depiction of actions, creating a natural flow between individual activities. This level of detail not only enhances the linguistic diversity of captions but also improves their utility in datasets for tasks such as human action understanding and multimodal learning. 

\subsubsection{Human Checking via Physical Knowledge and Logical Coherence.}
\label{sec:hd-checking-phy}
To reduce discrepancy and hallucination in LLM-generated captions, we implement a human checking verification stage that enforces physical plausibility, scene logic, and dataset conventions before acceptance as ground truth. Four graduate-level raters conduct this review using a structured checklist and edit protocol.

We validate captions against the following criteria:
\emph{(1) Physical feasibility and kinematics.} Body-pose and object-state transitions must obey continuity. For example, ``cup is empty'' $\rightarrow$ ``pouring into cup'' $\rightarrow$ ``cup becomes full,'' not the reverse.
\emph{(2) Scene and environment consistency.} Actions and objects must be compatible with the room and floor plan (e.g., ``brushing teeth'' in a bathroom; “watching TV” requires a visible or plausibly placed TV). Captions must not assert observations outside a sensor’s FOV.
\emph{(3) Temporal and causal coherence.} Event order must be logically progressive (e.g., ``grabbing utensil'' precedes ``eating'').
\emph{(4) Affordance and commonsense constraints.} Interactions must respect object affordances (e.g., ``stirring with a fork/spoon,'' not ``stirring with a phone''). A caption may cover multiple actions, but each action span must be temporally localizable. Note that in \workname, captions prioritize action/scene semantics over appearance details that are modality-incongruent. This human verification serves as a reliability gate, yielding captions that are (i) physically plausible within the recorded environments, (ii) temporally and causally coherent, and (iii) aligned with the action ontology.

\subsection{Put All Things Together}
\label{sec:DCP}

\subsubsection{Hardware and Environment Setup.}
\begin{figure*}[t]
    \centering
    \begin{subfigure}[t]{.56\textwidth}
        \centering
    \includegraphics[width=\linewidth]{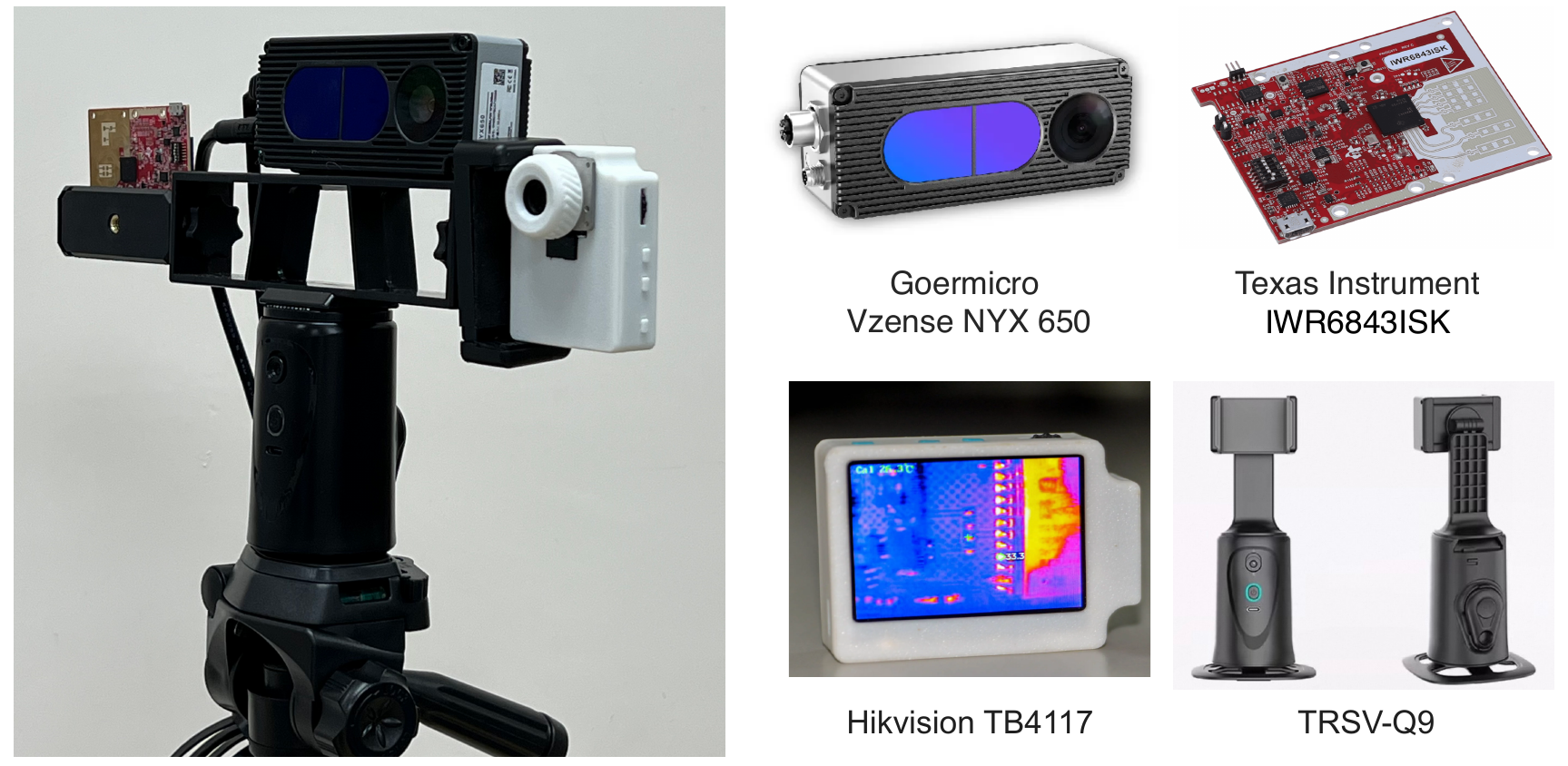}
    \caption{Ambient sensors include the Vzense NYX 650 for image sensing, the Texas Instruments IWR6843ISK for radar sensing, and the Hikvision TB4117 for thermal imaging. TSRV-Q9 is an AI tracking gimbal.}
    \label{fig:ambient_1}
    \end{subfigure}
    \hfill
    \begin{subfigure}[t]{.42\textwidth}
        \centering
    \includegraphics[width=\linewidth]{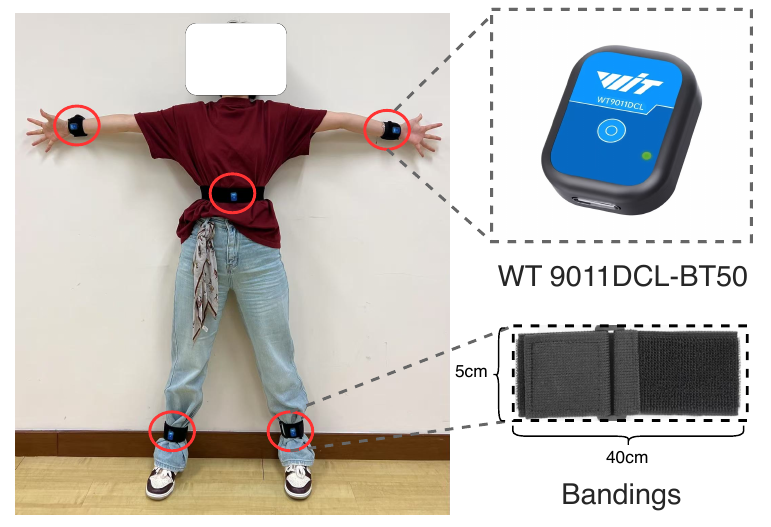}
    \caption{Wearable sensor includes 5 IMU sensor placements on the wrists, ankles, and waist using bandings, and the WitMotion WT 9011DCL-BT50 IMU module.}
    \label{fig:imu_1}
    \end{subfigure}
    \vspace{-10pt}
    \caption{Photos of our ambient and wearable sensor hardware.}
    \vspace{-10pt}

    \label{fig_sensor_1}
\end{figure*}

In this section,  we describe our hardware and environment configurations. As shown in Fig.~\ref{fig:ambient_1}, firstly, we use a Goermicro Vzense NYX 650 camera to capture RGB, depth, and infrared data.  Next, we use a Texas Instruments IWR6843ISK mmWave radar operating. In addition, we use a Hikvision TB4117 thermal imaging camera for precise temperature measurement. Moreover, we use a TSRV-Q9 AI Tracking Gimbal which is designed for precise automatic tracking and stabilization. In practice, we fix the sensor’s angle and position during data collection. Lastly, we use the Bluetooth 5.0-enabled WitMotion WT9011DCL-BT50 as our 9-axis Inertial Measurement Unit (IMU) for precise tracking of acceleration, angular velocity, and magnetic field.
Each participant was equipped with 5 of these devices, with sensors placed on the wrists, ankles, and waist using adjustable bands, shown in Fig.~\ref{fig:imu_1}. We collected data from two indoor environments, with a focus on four common room settings: the living room, kitchen, bedroom, and bathroom.
Our environmental setup not only enables fine-grained monitoring of human activities but also supports the integration and analysis of data across multiple modalities.\footnote{We provide more details of hardware and environment in \S \ref{sec:appx:hard_details} and \S\ref{sec:appendix:env}.}

\subsubsection{Demographic Characteristics  of Participants}
We recruited 30 participants (40\% male, 60\% female) with an age range of 20-23 years. BMI ranged from 16.41 to 29.02, with a mean of 24.54. Additionally, we collected data on participants' activity habits, indicating an average session duration of approximately 22.67 minutes and an average exercise frequency of 1.7 times per week, where low, moderate, and high intensities are assigned scores of 1, 2, and 3, respectively. These metrics suggest a relatively balanced distribution of height and weight among participants and highlight their tendency toward low-frequency, short-duration exercise routines. This dataset provides a meaningful baseline for the development and evaluation of computational models or systems aimed at activity recognition and health monitoring, ensuring both generalizability and reliability in human-centric data.

\subsubsection{Data Synchronization and Annotation}
To ensure precise alignment across all modalities, we adopt the global time from the host computer as the reference for synchronization. We use a marker, i.e., a director's board, to define the start and end points of the alignment process, enabling consistent temporal boundaries for all recorded data. RGB data serves as the primary modality for synchronization due to its high temporal resolution and consistency. Radar and IMU data are recorded with timestamps rigorously aligned to the global time, ensuring that all data streams are temporally synchronized to a high degree of accuracy.

For caption data annotation, captions are pre-generated (refer to \S\ref{sec:caption_generation} for more details), and subsequently used during data collection. This process ensures that the descriptions of each video segment are naturally aligned with the corresponding actions, minimizing annotation errors. In addition to caption-level alignment, individual action annotations are performed with meticulous care. Each video segment is manually labeled and segmented on a frame-by-frame basis to achieve the highest possible precision. Special attention is given to segment transitions and ambiguous actions to avoid misalignment or mislabeling, which can significantly impact downstream tasks. This manual process provides an accurate foundation for training and evaluating computational models. We provide more details of data statistics and data visualizations in \S\ref{sec:data_stat} and \S\ref{sec:appx:dv}, respectively.

\section{Experimental Setup}
Here, we describe our tasks, baseline, metrics, and implementation details. Note that in this paper, LLMs are used for generality without distinguishing between modalities.

\subsection{Tasks Descriptions}
\subsubsection{{HAR Task}} HAR is a task focused on identifying and classifying human activities such as walking, running, sitting, and standing from sensor data.  We define 40 classes across seven categories for recognition as our HAR tasks.

\subsubsection{{HAU Tasks}} HAU goes beyond basic HAR by capturing richer semantic information. Unlike HAR, which focuses on predefined actions, HAU seeks to understand the context of action sequences, including spatiotemporal semantics, relationships between actions, their order, and interactions with objects or the environment.
In particular, we define four sub-tasks in HAU as follows:  

    \begin{itemize}[leftmargin=*]
        \item \textbf{Caption Comparison:} This sub-task involves comparing the captions generated by the LLMs with the ground truth captions to evaluate the LLMs' capability for accurate description generation.  

        \item \textbf{{Context Analysis:}} This task requires that the LLMs must identify the correct context exhibited by the participants. In particular, we hope LLMs can recognize when actions are performed in a relaxed, calm, or hurried manner.
        
        \item \textbf{Sequential Action Reordering:} The model observes data containing actions in a shuffled order and accurately reorders them into the correct sequence.
        
        \item \textbf{Action Selection:} The LLMs observe the data to select the correct actions from a predefined pool of 40 actions.

    \end{itemize}

\subsubsection{{\HARn~Task}} \HARn~goes beyond HAU's semantic understanding by adding reasoning capabilities to infer intentions, causal relationships, and logical action sequences, which involves predicting outcomes. Specifically, the model must predict the next action from a provided list based on a series of preceding actions.

\subsection{Baseline and Metrics}  
\subsubsection{HAR Task}
 We use ResNet-50~\cite{he2015delving} for its visual recognition effectiveness. Radar data is processed with PointNet~\cite{qi2017pointnet}, enhanced by feature engineering to capture spatial characteristics. Skeleton data employs MotionBert~\cite{zhu2023motionbert}, using a dual-stream transformer and a multilayer perceptron, with 17 3D joints extracted via Human3.6M-compliant pose estimation models~\cite{mmpose2020}. IMU data is handled by a 1D-CNN~\cite{tang2020omni} with three convolutional layers and transformer encoder layers~\cite{vaswani2017attention} followed by a linear classification head. HAR task evaluation uses Accuracy, Precision, Recall, and F1-score.

\subsubsection{HAU and HARn Tasks.} 
We evaluate these two tasks via the latest video LLMs and visual reasoning LLMs, including InternVL2.5-2B (InternVL-2B)~\cite{chen2024internvl}, InternVL2.5-8B (InternVL-8B)~\cite{chen2024internvl}, QwenVL2.5-3B/7B (QwenVL-3B/7B)~\cite{Qwen2.5-VL}, VideoLLaVA-7B (VLLaVA-7B)~\cite{lin2023video}, and VChatR1-7B~\cite{li2023videochat}. For the caption comparison task in HAU, we use metrics including BERTScore (F1-Score), ROUGE-1, ROUGE-L, BLEU-1, and METEOR scores. Additionally, we use accuracy to evaluate the emotion analysis, sequential action reordering, activity selection, and \HARn~task. 
 
\subsection{Implementation Details}
\label{sec:appendix:implementation}
For both training schemes, we set the learning rate to 0.001, use a batch size of 64, and update parameters with the Adam~\cite{kingma2014adam} optimizer. For HAR tasks, we randomly split 80\% as the training set and 20\% as the testing set in all modalities. 
For RGB, IR, thermal, and depth modalities, models are initialized with ResNet-50 pre-trained weights, while for IMU and radar modalities, baseline models are initialized using Kaiming initialization~\cite{he2015delving} since no general pre-trained models are available for these modalities. Besides, for the skeleton modality, we initialize MotionBERT~\cite{zhu2023motionbert} as our backbone with weights pre-trained on the NTU RGB+D dataset~\cite{shahroudy2016ntu}. We provide the process details of depth data in \S\ref{sec:appx:depth_data}.

All video LLMs are evaluated under a zero-shot paradigm using their default configurations and task-specific system prompts.We directly use the whole video clip as our input.
In HAU tasks, models receive different prompts: (1) captioning-\textit{``Describe what the person in the video is doing. You can briefly mention the background or setting, but focus mainly on understanding the person's actions.''}; (2) emotion analysis-\textit{``What emotion does the person experience while performing the activities?''}; (3) sequential action reordering-\textit{``What activity is the person performing in the video? You must choose only from the following activities: \{Class Set\}. You can choose multiple activities if necessary.''}; (4) action selection-\textit{``Please sort the following activity lists in chronological order based on the video content.''.}; (5) HARn-\textit{``What activity is the person likely to do next?''.}

\section{Benchmarks}
We present three benchmarks in \workname: HAR, HAU, and HARn. For HAU and HARn we report results for vision only. The main reason is that we have strong and widely used LLMs exist for RGB, for example QwenVL, while comparable models for sensors such as IMU or mmWave are not yet available. Note that a direct comparison across these very different modalities would not be fair. Our results are meant to show that each modality contains useful task relevant information, not to rank the modalities or claim that one is the best.

\subsection{Benchmark of HAR}
To verify that the \workname~dataset contains sufficiently new knowledge for the HAR tasks in different data modalities, we provide a benchmark for the HAR task. The remaining content is structured around addressing the following two questions: \textit{(1) Does the dataset contain valuable knowledge?} \textit{(2) What are the challenges in this task?}

\subsubsection{Overall Performance}
As shown in Table~\ref{tab:cross_trial}, performance varies by modality, with an overall accuracy of 76.52\%. With standard supervised training, vision-based inputs work best. Thermal gives the top results, with accuracy 92.57\% and F1 score 93.36\%. Depth and RGB follow, with F1 scores of 90.93\% and 91.28\%. IR is lower than the other visual channels but still strong, with F1 90.46\%. Skeleton features also perform well, with accuracy 79.08\% and F1 84.17\%. In contrast, other sensors are much lower: mmWave reaches 46.63\% accuracy and 44.53\% F1, and IMU reaches 45.52\% accuracy and 38.32\% F1. These gaps likely come from the lower spatial resolution and signal-to-noise ratio of mmWave, the sensitivity of IMU to placement and orientation, and larger shifts across trials for these sensors. Overall, the results show that the dataset provides useful information for HAR.

\begin{table}[t]\small
    \centering
    \caption{Overall Performance of HAR Task in \workname.}
    \vspace{-10pt}
    \begin{tabular}{ccccc}
        \hline
        \textbf{Modality} & \textbf{Accuracy} & \textbf{Precision} & \textbf{Recall} & \textbf{F1-Score} \\
        \hline
        RGB      & 90.89\% & 92.24\% & 91.02\% & 91.28\% \\
        Depth    & 90.46\% & 91.76\% & 90.75\% & 90.93\% \\
        IR       & 90.22\% & 91.53\% & 89.94\% & 90.46\% \\
        Thermal  & 92.57\% & 93.54\% & 93.50\% & 93.36\% \\
        mmWave   & 46.63\% & 48.29\% & 46.63\% & 44.53\% \\
        IMU      & 45.52\% & 40.84\% & 38.00\% & 38.32\% \\
        Skeleton & 79.08\% & 91.46\% & 79.08\% & 84.17\% \\
        \hline
    \end{tabular}
    \vspace{-10pt}
    \label{tab:cross_trial}
\end{table}
 
\subsubsection{Long-tailed Class Performance}
As shown in Fig.\ref{fig:action-fre-2}, \workname exhibits a long-tailed class-frequency distribution, which may affect the results. The imbalance ratio is approximately 10 (i.e., the most frequent class appears about ten times as often as the rarest), indicating a moderate, though not extreme, level of imbalance~\cite{shuai2022balancefl}. This issue can be alleviated with standard techniques such as data resampling~\cite{ye2023licam}, data augmentation\cite{jiang2022pgada,jiang2023dual}, or balanced-loss objectives~\cite{shuai2022balancefl}. In particular, as shown in Fig.~\ref{fig:exp:resampling}, applying class-balanced resampling on RGB, IMU and Skeleton modalities in \workname~yields a measurable improvement in accuracy.
In particular, resampling consistently improves accuracy across modalities, with the largest gain for RGB from 90.89\% to 96.16\%, and smaller but noticeable gains for IMU.

\subsubsection{Cross-subject Performance}
\label{sec:exp:cross-subject}
\begin{figure}[t]
    \centering
    \begin{subfigure}[b]{0.26\textwidth} 
        \centering
        \includegraphics[width=.95\textwidth]{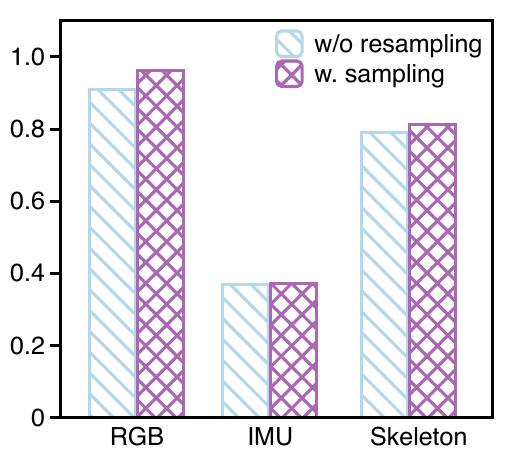} 
        \vspace{-5pt}
        \caption{Long-tailed Class.}
        \label{fig:exp:resampling}
    \end{subfigure}
    \hfill
    \begin{subfigure}[b]{0.19\textwidth} 
        \centering
        \includegraphics[width=.95\textwidth]{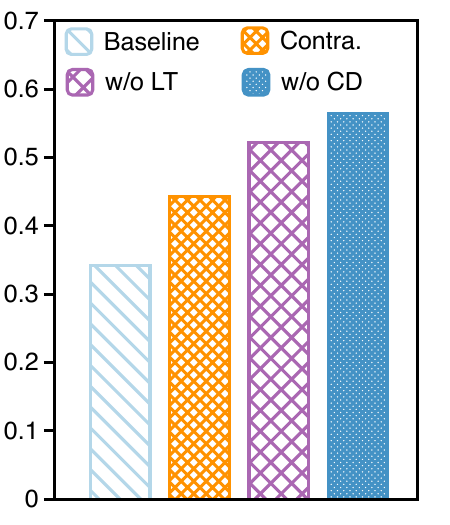} 
        \vspace{-5pt}
        \caption{Cross-subject.}
        \label{fig:exp:cross-subject}
    \end{subfigure}
    \vspace{-8pt}
    \caption[]{Long-tail (RGB, IMU, Skeleton) and cross-subject (RGB) performance. \textit{w/o LT} means long‑tail classes are removed; \textit{Contra.} means contrastive learning is added; \textit{w/o CD} means cross‑domain data are excluded.}
    \label{fig:motivation:limit_LLMs}
    \vspace{-10pt}
\end{figure}

We evaluate cross-subject performance of \workname using a leave-one-subject-out (LOSO) protocol. In each fold, one subject is held out for testing and the remaining subjects are used for training; results are averaged over five folds. As shown in Fig.~\ref{fig:exp:cross-subject}, the Baseline (LOSO on RGB only) exhibits a substantial accuracy drop due to subject shift and the long-tailed label distribution. Performance improves as we progressively mitigate these factors: removing long-tailed classes (w/o LT) yields a clear gain; adding contrastive learning (Contra.) further strengthens subject-invariant representations; and excluding cross-domain data (w/o CD) achieves the best result by eliminating domain shift, i.e., reaching $56.38\%$. Note that in \workname, CD denotes that we use training and testing data in the same physical environment. Compared with the performance of conventional HAR, accuracy drops markedly in the cross-domain setting, which is intrinsically challenging due to domain shift; even state-of-the-art methods report only
$\sim60\%$ accuracy in this setup~\cite{liu2024recovering}.

\subsection{Benchmark of HAU}
In HAU benchmark, our goal is to benchmark the task performance of different models and different modalities of the following four sub-tasks.
\subsubsection{Results of Caption Comparison} 

As shown in Table~\ref{tab:benchmark_hau_task1}, different models excel on different metrics and modalities. VLLaVA‑7B achieves the best BERTScore F1 and the highest ROUGE‑1/ROUGE‑L on the depth and thermal modalities. QwenVL‑3B attains the top ROUGE and BLEU‑1 scores on RGB and IR and remains competitive elsewhere. VR1Chat‑7B frequently yields the best BLEU‑1 and METEOR (e.g., RGB and IR), indicating strong fluency. In contrast, InternVL‑2B/8B obtain decent BERTScore F1 but lag behind on ROUGE and BLEU. Overall, higher‑capacity models (e.g., 7B) tend to outperform the InternVL‑2B/8B baselines, although the 3B Qwen model is a notable exception that matches or surpasses some 7B models on several metrics.

\begin{table}[t]\small
    \centering
    \caption{Results of caption comparison. We report BERTScore F1 (B.-F1), ROUGE-1 (R.-1), ROUGE-L (R.-L), BLEU-1 (B.-1), and METEOR (MET.).}
    \vspace{-10pt}
    \resizebox{.95\columnwidth}{!}{
    \begin{tabular}{lccccc}
        \toprule
        \textbf{Model} & \textbf{B.-F1} & \textbf{R.-1} & \textbf{R.-L} & \textbf{B.-1} & \textbf{MET.} \\
        \hline
        \multicolumn{6}{c}{\textbf{RGB}} \\
        \hline
        InternVL-2B  & 84.39\% & 4.33\%  & 3.64\%  & 0.61\%  & 3.97\% \\
        InternVL-8B  & 84.07\% & 3.04\%  & 2.53\%  & 0.72\%  & 3.63\% \\
        QwenVL-3B    & 86.22\% & 18.40\% & 13.80\% & 21.46\% & 19.89\% \\
        QwenVL-7B    & 85.47\% & 14.79\% & 12.05\% & 18.04\% & 22.21\% \\
        VLLaVA-7B    & 86.40\% & 16.12\% & 12.77\% & 12.86\% & 11.58\% \\
        VR1Chat-7B   & 86.24\% & 17.42\% & 13.66\% & 21.62\% & 23.18\% \\
        \hline
        \multicolumn{6}{c}{\textbf{Depth}} \\
        \hline
        InternVL-2B  & 84.09\% & 4.63\%  & 3.95\%  & 0.53\%  & 3.67\% \\
        InternVL-8B  & 83.95\% & 2.89\%  & 2.35\%  & 0.73\%  & 3.52\% \\
        QwenVL-3B    & 85.03\% & 15.00\% & 10.76\% & 18.58\% & 16.89\% \\
        QwenVL-7B    & 84.55\% & 12.70\% & 10.65\% & 16.64\% & 18.98\% \\
        VLLaVA-7B    & 85.94\% & 16.31\% & 13.53\% & 12.07\% & 10.37\% \\
        VR1Chat-7B   & 84.69\% & 14.17\% & 11.49\% & 17.73\% & 19.39\% \\
        \hline
        \multicolumn{6}{c}{\textbf{IR}} \\
        \hline
        InternVL-2B  & 84.24\% & 4.38\%  & 3.68\%  & 0.58\%  & 4.09\% \\
        InternVL-8B  & 84.22\% & 3.05\%  & 2.54\%  & 0.64\%  & 3.68\% \\
        QwenVL-3B    & 86.49\% & 18.56\% & 13.99\% & 22.17\% & 18.85\% \\
        QwenVL-7B    & 85.38\% & 14.71\% & 11.93\% & 18.25\% & 21.50\% \\
        VLLaVA-7B    & 86.25\% & 15.64\% & 12.81\% & 12.25\% & 10.78\% \\
        VR1Chat-7B   & 86.03\% & 16.51\% & 12.96\% & 21.23\% & 22.05\% \\
        \hline
        \multicolumn{6}{c}{\textbf{Thermal}} \\
        \hline
        InternVL-2B  & 84.30\% & 5.05\%  & 4.28\%  & 0.38\%  & 3.93\% \\
        InternVL-8B  & 83.85\% & 2.74\%  & 2.24\%  & 0.73\%  & 3.37\% \\
        QwenVL-3B    & 85.04\% & 14.78\% & 11.40\% & 18.13\% & 16.95\% \\
        QwenVL-7B    & 84.48\% & 12.44\% & 10.31\% & 15.41\% & 20.24\% \\
        VLLaVA-7B    & 85.85\% & 16.24\% & 13.32\% & 11.31\% & 10.16\% \\
        VR1Chat-7B   & 84.94\% & 14.23\% & 11.51\% & 17.90\% & 19.86\% \\
        \bottomrule
    \end{tabular}
    }
    \vspace{-10pt}
    \label{tab:benchmark_hau_task1}
\end{table}

\subsubsection{Results of Context Analysis}
\begin{figure*}[t]
    \centering
    \begin{subfigure}[b]{0.45\textwidth} 
        \centering
        \includegraphics[width=\textwidth]{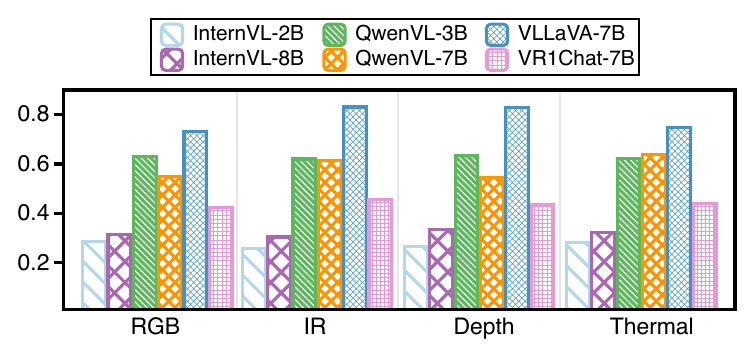} 
        \vspace{-20pt}
        \caption{Context Analysis}
        \label{fig:emotion_acc}
    \end{subfigure}
    \hfill
    \begin{subfigure}[b]{0.45\textwidth} 
        \centering
        \includegraphics[width=\textwidth]{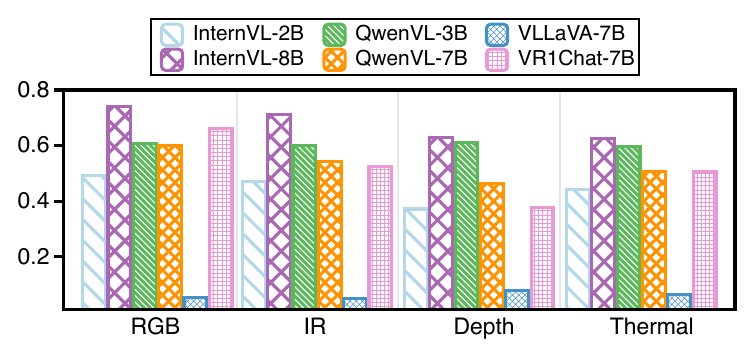} 
        \vspace{-20pt}
        \caption{Sequential Action Reordering}
        \label{fig:exp:seq}
    \end{subfigure}
    \vspace{-10pt}
    \caption[]{Accuracy of context analysis and sequential action reordering in \workname. }
    \label{fig:exp:HAU_task_2}
    \vspace{-10pt}
\end{figure*}

As shown in Fig.~\ref{fig:emotion_acc}, the average accuracy is $50.52\%$. In particular, VLLaVA‑7B delivers the best overall context accuracy across modalities, leading on RGB, IR, and Depth. QwenVL‑3B follows closely and remains competitive on Thermal. InternVL‑2B and InternVL‑8B stay at the lower end, with accuracies roughly between 24 and 35 percent across modalities, suggesting that although these architectures work well for general captioning, they are weaker at modeling contextual cues. VR1Chat‑7B achieves mid‑range results, typically around 42 to 50 percent, with slightly better scores on RGB and Depth than on IR and Thermal. The spread between models is largest on IR and Depth, where the best systems approach 80 percent while the weakest remain below 30 percent, underscoring the challenge of interpreting non‑RGB signals.

\subsubsection{Results of Sequential Action Reordering}
\label{sec:exp:SAR}

As shown in Fig.~\ref{fig:exp:seq}, the average accuracy is $47.24\%$ and no single model dominates across all modalities. We notice that InternVL‑8B achieves the best accuracy on RGB and IR, reaching roughly three quarters in both cases. QwenVL‑3B edges out the others on Depth, with scores in the mid‑sixties. VR1Chat‑7B is strongest on Thermal, around sixty percent. QwenVL‑7B is steady across modalities but does not lead any of them. VLLaVA‑7B performs poorly, i.e., close to chance on every modality, while InternVL‑2B also sits toward the lower end. Across modalities, IR and RGB are generally the most informative for this task, Depth is in the middle, and Thermal is the most challenging. These results suggest that architecture–modality compatibility matters: capacity helps in places, but the top results come from different models on different sensors.

\subsubsection{Results of Action Selection }
\label{sec:exp:AS}
\begin{figure}[t]
    \centering
    \includegraphics[width=.95\linewidth]{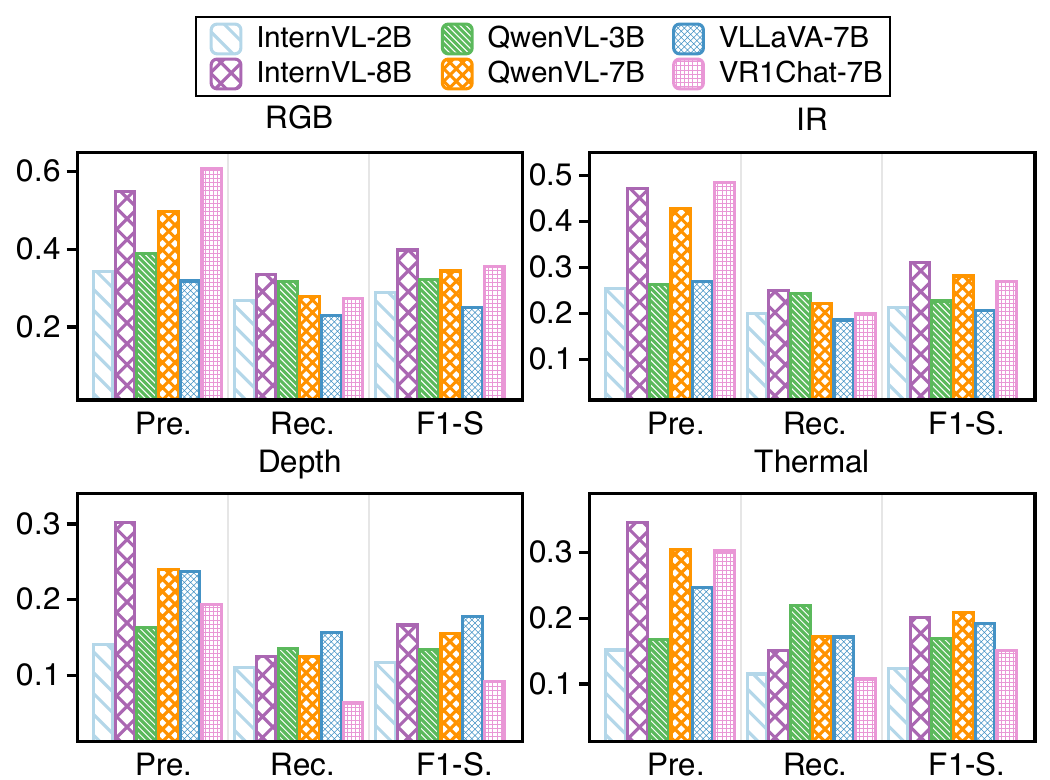}
    \vspace{-10pt}
    \caption{Accuracy of action selection tasks in HAU.}
    \vspace{-10pt}
    \label{fig:exp:action_select}
\end{figure}

As shown in Fig.~\ref{fig:exp:action_select}, the average accuracy is $24.54\%$. We notice that QwenVL‑7B generally attains the best or near‑best scores across all three metrics and modalities, indicating strong action‑selection capability, VLLava‑7B is consistently competitive, with notable strength on infrared and thermal inputs. InternVL‑8B delivers mid‑range results, with its strongest performance on RGB, while InternVL‑2B trails the other models, reflecting the limitations of a smaller‑capacity model. Across modalities, RGB and infrared typically outperform depth and thermal, suggesting that these signals provide more task‑relevant information. These findings underscore the joint importance of model scale and sensing modality for robust action selection.

\subsection{Benchmark of HARn}
We have evaluated the performance of LVLMs in inferring intentions and causal relationships in human action sequences.

\subsubsection{Results of \HARn}
\begin{figure}[tb]
    \centering
    \includegraphics[width=.95\linewidth]{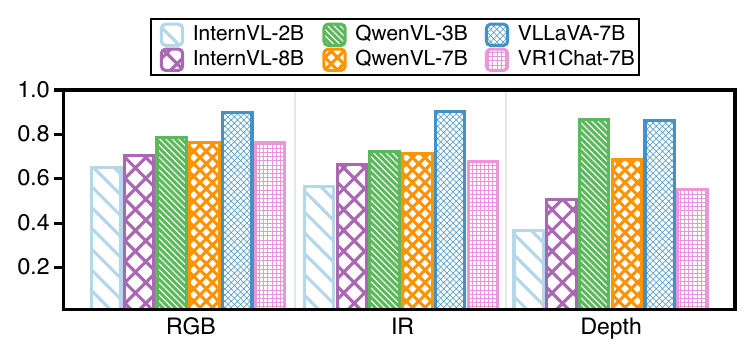}
    \vspace{-10pt}
    \caption{Results of \HARn.}
    \vspace{-10pt}
    \label{fig:exp:HARn}
\end{figure}

As shown in Fig.~\ref{fig:exp:HARn}, the average accuracy is $70.25\%$. We notice that VLLava‑7B leads on RGB and infrared and remains competitive on depth. QwenVL‑7B performs consistently well across all modalities and is close to the top overall. QwenVL‑3B stands out on depth, surpassing several larger models. We found that InternVL‑8B and VR1Chat‑7B deliver mid‑range results, while InternVL‑2B trails the others. On average, depth and infrared yield higher accuracy than RGB, indicating that these signals carry more information for \HARn. These findings highlight the combined impact of model capacity and sensing modality on robust human‑activity reasoning.

\subsubsection{Why reasoning model works in \HARn?} 
\begin{figure}[t]
    \centering
    \includegraphics[width=.95\linewidth]{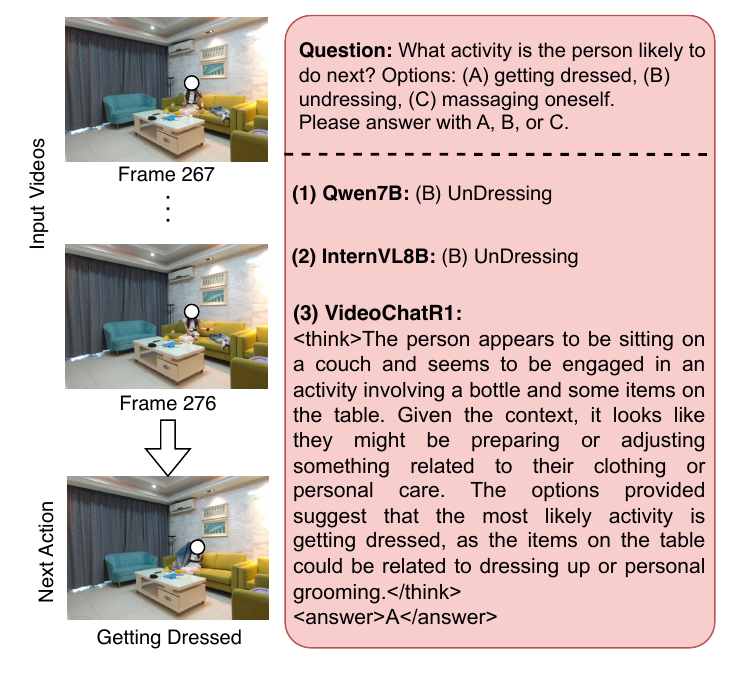}
    \vspace{-10pt}
    \caption{An illustration  of analysis why the reasoning model performs well than the captioning model.}
    \vspace{-10pt}
    \label{fig:exp:HARn_re}
\end{figure}

Fig.~\ref{fig:exp:HARn_re} illustrates the superiority of reasoning-based models in \HARn~tasks. Unlike captioning models, e.g., Qwen-7B, InternVL-8B, that misinterpret superficial cues, the reasoning model, i.e., VR1Chat, leverages contextual understanding and logical inference. It associates observed actions, such as interacting with items on the table, with the most likely next action (i.e., ``Getting Dressed''). Additionally, the reasoning model excels in handling ambiguity and provides transparent explanations, enhancing interpretability. This capability to integrate temporal reasoning and contextual synthesis makes reasoning models more reliable for HARn tasks, where understanding intent and action progression is critical.

    \section{Discussions }
\label{sec:discussion}

Due to the page limitation, we have provided more discussion of \workname{} in our Appendix~\S\ref{sec:appendix:more_dis}.

\subsubsection*{Limitations}
\workname{} is a controlled dataset collected in two indoor environments with 30 participants with ages from 20 to 23. As such, it lacks population‑level and ecological diversity, and generalization to other settings, long‑horizon activities, or populations with different motor patterns, such as children, older adults, the individuals with mobility impairments, is uncertain. Because actions are elicited from scene‑level captions, instruction‑following may shift motion distributions relative to naturalistic behavior despite preliminary, small‑scale evidence of caption similarity. In future releases, we may refine our adjudication protocol, such as reporting inter‑rater reliability for the initial independent annotations, which can provide a more comprehensive evaluation.

\subsubsection*{Future Directions}
\workname{} can scale along two axes: (1) adding interaction actions among participants and longer routines; and (2) augmenting the current seven modalities, with complementary signals such as audio, tactile/contact sensing, and lightweight physiology (heart rate, EEG). We also plan to broaden the participant pool and environments to strengthen generalizability. In parallel, \workname{} serves as a standard benchmark for HAR, and as a testbed for LLM‑based action understanding and reasoning. Its tightly synchronized multimodal streams make it a practical educational resource for teaching sensor fusion, and multimodal reasoning.

    \section{Related Works}

\subsubsection*{\textbf{Human Action Recognition Datasets}}
HAR analyzes and classifies human actions using various sensors. Vision-based datasets mainly utilize RGB and RGB-D data to capture activities. For example, NTU-60~\cite{shahroudy2016ntu} provides 56,880 videos of daily and health-related actions, while UTD~\cite{chen2015utd} records 27 activities from 8 subjects for classification. Similarly, PKU-MMD~\cite{liu2017pku} and NTU-120~\cite{liu2019ntu} leverage RGB, depth, and skeleton data, supporting 66 and 120 actions, respectively. Sensor-based datasets use wearable or environmental sensors like IMUs, gyroscopes, or radar. UMAFall~\cite{casilari2017umafall} employs IMU sensors on the chest, waist, wrist, and ankle for fall detection, while Epic-Kitchen~\cite{damen2022rescaling} combines IMU, RGB, and optical flow to analyze over 90,000 action segments in kitchen environments. Smaller datasets, such as USC~\cite{zhang2012usc}, Shoaib~\cite{shoaib2014fusion}, HHAR~\cite{stisen2015smart}, and UCI~\cite{reyes2016transition}, focus on common activities like walking using IMU data. Radar-based datasets, such as HuPR~\cite{Lee_2023_WACV}, integrate radar and RGB for privacy-preserving action recognition. The emerging dataset Thermal-IM~\cite{ThermalIM2023} employs thermal imaging and multimodal data to address challenges such as lighting variations and occlusion, enabling effective long-term tracking. MM-Fi~\cite{yang2023mm} integrates RGB, depth, and radar, offering over 320,000 samples for 27 activities conducted by 40 subjects. While limited in scale, mRI~\cite{an2022mri} also combines IMU, covering 12 activities performed by 20 subjects. However, existing datasets lack comprehensive HAR data from diverse IoT devices. 

\subsubsection*{\textbf{Human Action Understanding Datasets}}
HAU involves comprehending actions through perceptual, contextual, and experiential integration, covering recognition, intention, and narrative understanding. Multimodal datasets such as PKU-MMD~\cite{liu2017pku} and Ego-Exo4D~\cite{grauman2024ego} support the evaluation of algorithms for understanding complex activities. Captioning-based datasets such as Ego-4D~\cite{grauman2022ego4d} and Ego-Exo4D include 3,670 hours of videos with narrations to enrich activity understanding, while Tarsier2~\cite{yuan2025tarsier2advancinglargevisionlanguage} uses large language models for detailed descriptions. Reasoning-based datasets like ActivityNet-QA~\cite{yu2019activitynet} and Next-QA~\cite{xiao2021next} focus on spatial, temporal and causal reasoning, with annotated question-answer pairs to enhance deeper video content understanding. DailySTR~\cite{qiu2024dailystr} further leverages the VirtualHome-AIST simulator to create a video-based dataset comprising a total of 80,573 question-answer (QA) pairs. However, these datasets often focus on single modalities or involve high annotation costs. \workname{} addresses these gaps as the first multimodal dataset for HAU, integrating understanding and reasoning across modalities to advance human action comprehension.


    \section{Conclusion}
In this paper, we present \workname, a large-scale multimodal dataset and benchmark in HAR, HAU and HARn with offering 58,445 samples across seven modalities and two environments. 
With three carefully designed benchmarks encompassing six tasks, our results demonstrate the robustness of \workname~in validating state-of-the-art models across three tasks.

\clearpage
    \bibliographystyle{ACM-Reference-Format}
    \bibliography{acmart}

\appendix

\section*{Appendix}

\section{Overview and Details}
\subsection{Overview of \workname}
\label{sec:appendix:overview-cuhk}
We introduce \workname, a large‑scale multimodal dataset designed to advance human action recognition (HAR), understanding (HAU), and next‑action reasoning (HARn). As shown on the left of the figure, we first build a Scene‑based Caption Generation framework to avoid spatiotemporal inconsistency and provide precise ground truth. Guided by ATUS, we define seven action themes and select 40 representative actions with reference to HHAR, UCI, and Cosmo. Large language models then compose coherent captions that stitch these actions into everyday scenes—living room, kitchen, bedroom, and bathroom—augmented with emotional styles such as relaxed or hurried. Using these captions as ground truth (center), 30 participants enact the scenes in two indoor environments, yielding over 58,445 samples across seven synchronized modalities: RGB, depth, thermal, infrared, skeleton, IMU, and mmWave. The setup includes a Vzense NYX 650 (depth), TI IWR6843ISK (mmWave), a Hikvision TB4117 (thermal), and five WitMotion WT9011DCL‑BT50 IMUs. Each recording is paired with its caption, forming rich data–caption pairs. On the right, \workname supports three benchmarks implemented as eight tasks. HAR performs single‑label classification. HAU assesses caption comparison with ground truth, context analysis (e.g., quickly, smoothly, calmly), sequential action reordering to test temporal reasoning, and action selection from a predefined set. HARn predicts the next action from textual descriptions. We evaluate across all modalities using SOTA models, InternVL2.5‑2B/8B, QwenVL2.5‑3B/7B (captioning) and VideoLLaVA‑7B, VideoChatR1‑7B (reasoning), and analyze long‑tail and cross‑subject effects.

\subsection{Motivation Details}
\label{sec:appendix:motivation}
As shown in Fig.~\ref{fig:motivation:limit_LLMs_1}, we provide an illustration of the limitations in current LLMs, such as Tarsier~\cite{wang2024tarsier} and Tarsier2~\cite{yuan2025tarsier2}, which face challenges in achieving accurate HAU with depth and thermal modalities, while RGB performs accurate results. We observed that the SOTA captioning models often make the following mistakes: providing inaccurate action descriptions, missing actions, or sometimes both, as shown in Fig.~\ref{fig:motivation:limit_LLMs_1}. The main reason is that these models are not trained on this modality or designed for the HAU task.  Thus, it is necessary to provide datasets with multimodal synchronized ``$\langle$data, caption$\rangle$'' pairs to enable models to understand such information effectively. 

\subsection{Hardware Details}
\label{sec:appx:hard_details}
\subsubsection{{Ambient sensors setup}} 

As shown in Fig.~\ref{fig:ambient_1}, firstly, we use a Goermicro Vzense NYX 650 camera to capture RGB, depth, and infrared data. Specifically, the Vzense NYX 650 cameras offer a 70° horizontal and 50° vertical field of view, operating at a frame rate of 10 frames per second. Leveraging 940nm infrared light, these cameras are well-suited for both indoor and outdoor environments, even under low-light or no-light conditions. Next, we use a Texas Instruments IWR6843ISK mmWave radar operating in the 60–64 GHz band. We configure it with a 20 fps frame rate, 0.044 m range resolution, a 5.03 m maximum unambiguous range, a 1.0 m/s maximum radial velocity, and a 0.13 m/s radial velocity resolution. This sensor excels in detecting objects, measuring distances, and tracking motion with high precision. In addition, we use a Hikvision TB4117 thermal imaging camera for precise temperature measurement. Featuring a $120 \times 160$-pixel resolution and compact $70 \times 46 \times 22.75$ mm dimensions, this device measures temperatures from $30^\circ \text{C}$ to $45^\circ \text{C}$ with 25 fps, making it ideal for thermal monitoring. Lastly, we use a TSRV-Q9 AI Tracking Gimbal, a compact (60 $\times$ 70 $\times$ 185 mm) and lightweight (220 g) device designed for precise automatic tracking and stabilization. Powered by a 3.7V/1200mAh battery, it supports 3.5 hours of continuous tracking. Compatible with devices up to 12 mm thick, it features 360° horizontal rotation and 180° manual vertical adjustment, making it ideal for dynamic content creation. In practice, we fix the sensor’s angle and position during data collection.

\subsubsection{{Wearable sensors setup}} We use the Bluetooth 5.0-enabled WitMotion WT9011DCL-BT50 as our Inertial Measurement Unit (IMU) for precise tracking of acceleration (\(\pm 16g\)), angular velocity (\(\pm 2000^\circ/s\)), and magnetic field (\(\pm 2~\text{Gauss}\)). It supports output frequencies ranging from 0.2 Hz to 200 Hz and provides angular measurements of up to \(\pm 180^\circ\) (X/Z) and \(\pm 90^\circ\) (Y). Powered by a 130 mAh battery, it delivers up to 40 hours of continuous operation with a maximum transmission range of 50 m in open space with 10 samples per second. 
Measuring\(32.5 \times 23.5 \times 11.6~\text{mm}\) in size and weighing just 9g, each participant was equipped with 5 of these devices, with sensors placed on the wrists, ankles, and waist using adjustable bands, shown in Fig.~\ref{fig:imu_1}.

\subsection{Environmental Details}
\label{sec:appendix:env}
We collected data from two indoor environments, with a focus on four common room settings: the living room, kitchen, bedroom, and bathroom. These rooms were selected to represent a diverse range of daily activities for studying and analyzing human behavior in realistic, everyday scenarios. To ensure systematic data collection, we documented all sensor locations in each room by marking their positions and taking photographic records. This meticulous annotation method provides thorough coverage of typical activities occurring within these spaces, facilitating robust data analysis. The floor plans, as depicted in Fig.~\ref{fig:room_vis}, offer detailed spatial representations of the two indoor environments, with icons highlighting the exact locations of the sensors deployed for data collection. Specifically, ambient sensors were strategically positioned to optimize coverage and data reliability. Furthermore, each photo depicts a room and serves to visually contextualize the sensor data, providing a visual context to the collected data. Our environmental setup not only enables fine-grained monitoring of human activities but also supports the integration and analysis of data across multiple modalities. 

\subsection{Processing Details of Depth Data.} 
\label{sec:appx:depth_data}
\begin{figure*}[t]
    \centering
    \includegraphics[width=\linewidth]{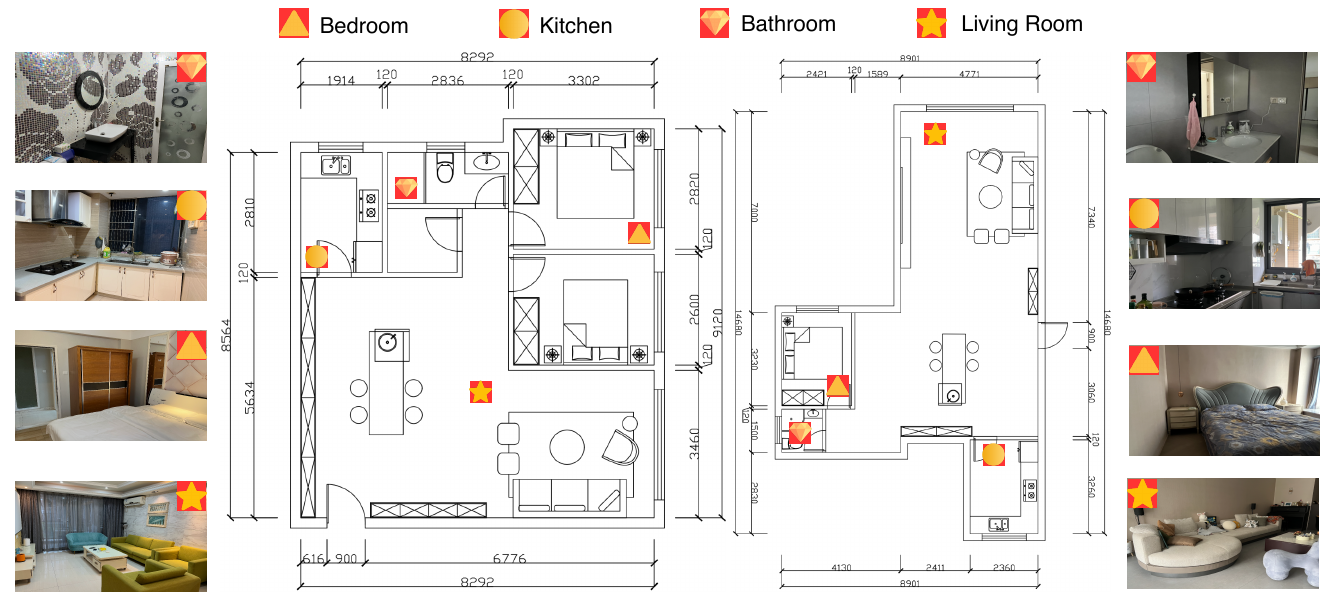}
    \vspace{-10pt}
    \caption{Environment Visualization (The left room is Room 1 and the right one is Room 2). Layout with room-wise visual annotations (Bedroom, Kitchen, Bathroom, and Living Room) showing corresponding example images and sensor placements. The icon indicates the location of the ambient sensor.}
    \vspace{-10pt}
    \label{fig:room_vis}
\end{figure*}

We observed that directly using this raw data fails to effectively capture the spatial information of depth since the raw depth data is 16-bit. To address this, we process the raw depth data via the two types of house floor plans, shown in Figure~\ref{fig:room_vis}. In particular, we filtered depth values outside a defined range, replacing them with zero to focus on relevant depth information for each specific environment. 
In Room 1 of Fig.~\ref{fig:room_vis}, the depth ranges are set as follows: Living Room [500, 5000], Kitchen [500, 3300], Bedroom [500, 3200], and Bathroom [500, 2800]. In Room 2 of Fig.~\ref{fig:room_vis}, the depth ranges are slightly different: Living Room [500, 4700], Kitchen [500, 3260], Bedroom [500, 3500], and Bathroom [500, 2000]. These ranges are tailored to accommodate the spatial characteristics of each scenario and room.

\begin{figure*}[t]
    \centering
    \includegraphics[width=\linewidth]{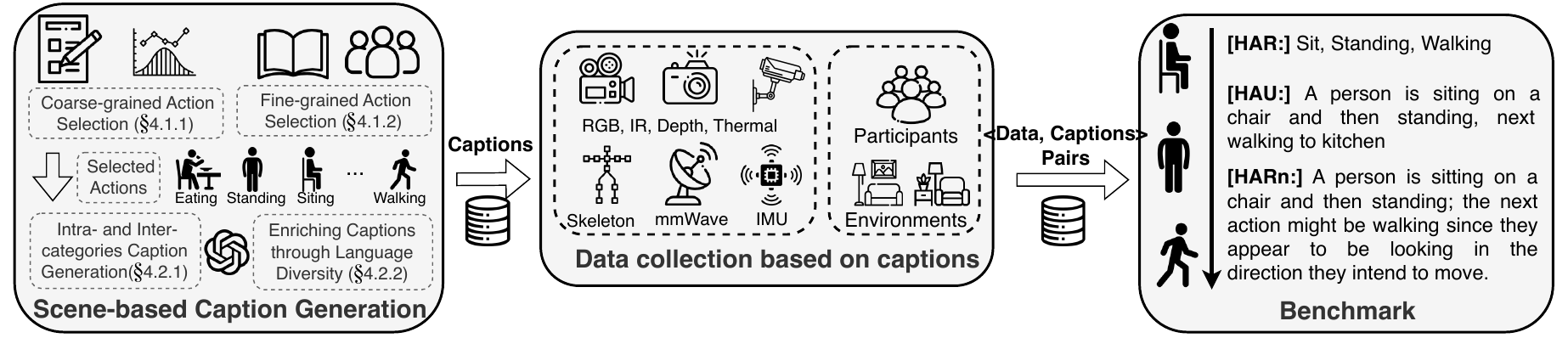}
    \vspace{-10pt}
    \caption{Overview of \workname. We first obtain a set of action categories based on coarse \& fine-grained action selection and then generate captions for these actions. We first create scenes based on selected actions and obtain captions. Then, we collect data from 30 participants across seven modalities and obtain the ``$\langle$data, caption$\rangle$'' pairs. Lastly, these data can support HAR, HAU, and HARn tasks. The distinction between these tasks lies in their objectives: HAR focuses on single-label classification, HAU generates detailed captions, and HARn predicts the next action within a spatiotemporal context.}
    \vspace{-10pt}
    \label{fig:ill}
\end{figure*}

\section{Data Description}

\subsection{\textbf{Data Statistics}}
\label{sec:data_stat}

\begin{figure*}[t]
    \centering
    \includegraphics[width=.96\linewidth]{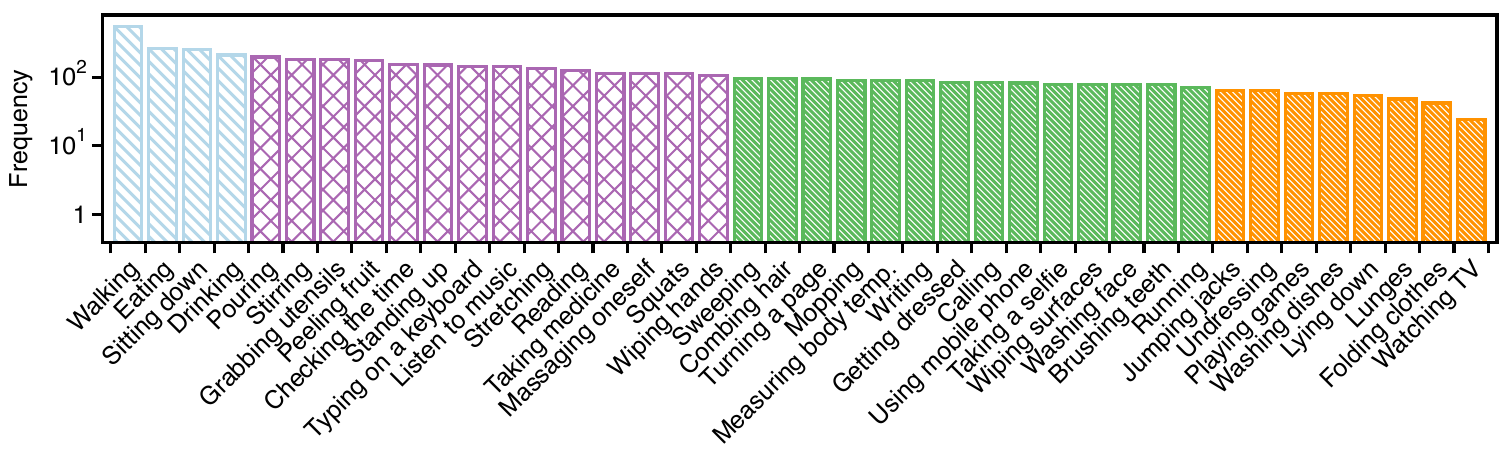}
    \vspace{-12pt}
    \caption{Data Statistics of \workname.}
    \vspace{-10pt}
    \label{fig:action-fre-2}
\end{figure*}

Here, we provide a statistical description of our dataset.  As shown in Fig.~\ref{fig:action-fre-2}, we show a clear frequency imbalance across human actions. High-frequency actions such as walking, eating, sitting down, and drinking water dominate the dataset, with occurrences exceeding 200, reflecting their ubiquity in daily life. Moderately frequent actions, including pouring a drink, stirring utensils, checking time, and standing up, appear between 50 and 150 times, indicating their importance in routine behaviors while being less universal. In contrast, actions such as folding clothes, watching TV, and playing a game are sparsely represented, with fewer than 20 occurrences, likely due to their specific or context-dependent nature. The dataset follows a long-tail distribution, where a small number of actions account for a large proportion of occurrences, while the majority are infrequent. This imbalance is a common characteristic of real-world datasets, which naturally prioritize capturing frequent, everyday behaviors. Despite this, the dataset spans a diverse range of categories, including basic daily activities, work-related tasks, household chores, and physical exercises, providing a rich foundation for human activity recognition. In particular, in \workname, each participant contributes over 30 minutes of footage with more than 100 samples. For example, vision modalities include 4,029 clips, with a total duration of 19 hours and 29 minutes.

\subsection{\textbf{Data Visualizations}}
\label{sec:appx:dv}
\begin{figure*}[t]
    \centering
    \includegraphics[width=0.95\linewidth]
    {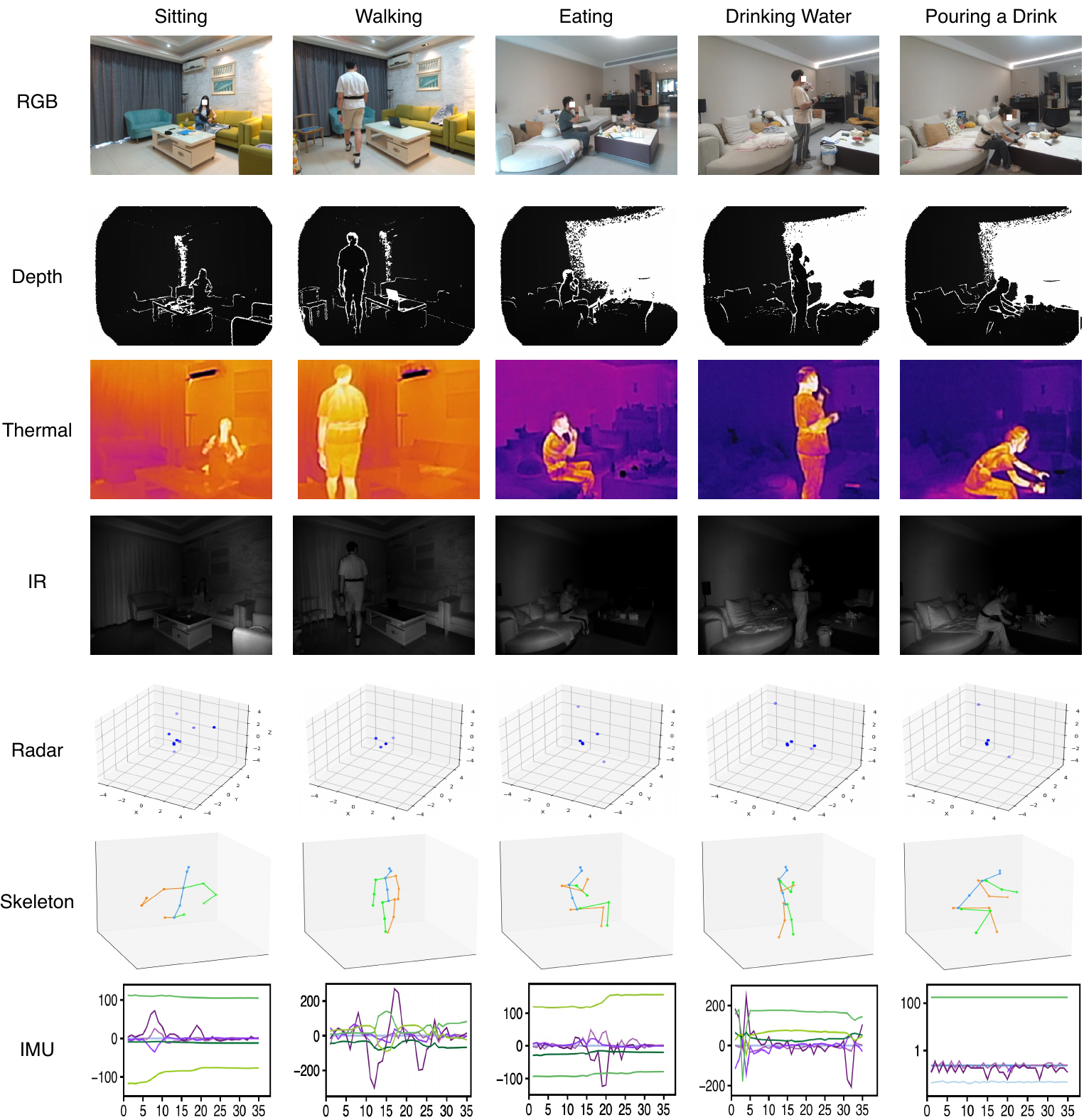}
    \vspace{-10pt}
    \caption{Visualization on ambient and wearable sensor data.}
    \label{fig:data_vis}
\end{figure*}

We also provide a visualization of our multimodal data. Fig.~\ref{fig:data_vis} illustrates the multimodal data from both ambient and wearable sensors. We present the visualization of common activities, including sitting, walking, eating, drinking water, and pouring a drink. The data includes RGB, Depth, Thermal, IR, Radar, 3D Skeletons, and IMU signals, representing a comprehensive set of modalities for activity recognition. In particular, RGB serves as the primary visual modality, capturing color and spatial context, which is essential for understanding the environment and participants' actions. Depth data provides spatial structure, highlighting the distance and geometry of objects and participants, while Thermal data visualizes temperature distributions, offering insights into heat signatures that may not be visible in other modalities. Infrared (IR) enhances visibility in low-light or dark environments, complementing RGB and Thermal data. Radar visualizes motion and spatial dynamics, making it particularly useful for detecting movement patterns. In addition, the 3D Skeletons, extracted using mmpose~\cite{mmpose2020}, provide key body joint positions and orientations, enabling precise pose estimation and body movement tracking. These skeletons are overlaid on RGB images for better interpretability of the captured actions. IMU data, collected from five body locations (right arm, left arm, waist, right leg, left leg), includes acceleration, angular velocity, and angles across the X, Y, and Z axes, offering fine-grained motion analysis. In \workname, each modality not only complements the others but is also capable of functioning independently.

\section{Discussion Details}
\label{sec:appendix:more_dis}

\subsection{Bias Checks of \workname} 
Because actions in \workname{} are collected from scene‑level captions, instruction‑following may shift motion distributions relative to naturalistic behavior. As preliminary evidence, we previously reported (\S\ref{sec:motivation:bias-GT}) a small four‑task study indicating high caption similarity. However,  caution is warranted when extrapolating to unconstrained, long‑horizon activities or to populations with different motor patterns (e.g., children, older adults, or individuals with mobility impairments).

\subsection{Scope Limitations}
\workname{} has controlled a dataset and benchmarks for multimodal perception, language grounding, and action reasoning in everyday indoor scenes. The current release covers 30 participants aged 20–23 and two indoor environments. Therefore, we do not claim population‑level or broad ecological diversity in \workname{}. Our main goal is to provide synchronized modalities (RGB/depth/IR/thermal/skeleton/IMU/mmWave) with scene‑level ground truth that stress‑test cross‑modal grounding and temporal reasoning. Expanding demographics and environments is planned future work.

\subsection{Verification Procedures}
In this release, we do not report inter‑rater reliability (IRR), e.g., Cohen’s $\kappa$ for binary checklist items or Krippendorff’s $\alpha$ for ordinal coherence ratings, because the final annotations are consensus labels produced after adjudication, making post‑hoc IRR on the gold set uninformative. Although the intended tasks (HAR, HAU, HARn) primarily target scene semantics and temporal dependencies, we view formal IRR as valuable for auditability; in future releases, we plan to report IRR on a double‑coded holdout and to release the rubric and adjudication protocol to facilitate external review.


\subsection{More Actions and Modalities.}
A key future direction for \workname~is addressing its scalability across actions and modalities. Firstly, while \workname~is comprehensive, it could be expanded to include more actions that involve interactions between multiple participants. In addition, while the current version includes data from seven modalities, future expansions could incorporate other modalities, such as audio, tactile sensors, heart rate, or EEG. These additional modalities would provide deeper insights into human actions by capturing complementary information, such as emotional states, physiological responses, or fine-grained tactile interactions, enriching the dataset's multimodal nature. Moreover, while \workname~currently includes data from 30 participants, expanding to a larger, more diverse pool of individuals is critical for improving generalizability. 

\subsection{More Discussions}
\subsubsection{Cross-subject and cross-domain HAR} Real deployments inevitably face domain shifts that go beyond subject identity. In \workname, shifts arise from (i) environment (different apartments and room layouts), (ii) sensor placements and view angles, (iii) lighting and thermal conditions, (iv) background clutter and occlusions, and (v) subject attire and execution styles. Although we have reported cross-subject degradation in \S\ref{sec:exp:cross-subject}, which indicates sensitivity to cross-domain, i.e., different physical environment factors, removing cross-domain data yields a best result of 56.56\% on RGB. This finding highlights that domain shifts remain a primary bottleneck even for vision-based HAR. Nevertheless, we believe it is feasible to improve model robustness for both cross-subject and cross-domain HAR.

\subsubsection{Discussion of zero-shot and fine-tuning.}
Our HAU and HARn evaluations intentionally use zero-shot LVLMs with task-specific prompts to expose modality gaps and avoid confounds from small-scale fine-tuning. This design choice surfaces two realities: (i) current LVLMs are primarily optimized for RGB, and (ii) reasoning over non-RGB modalities remains challenging without targeted adaptation. At the same time, our HAR results show that task-specific fine-tuning on CUHK-X materially improves recognition compared to off-the-shelf backbones (average 76.52\% across modalities), motivating a nuanced view of zero-shot vs fine-tuning. In this version, our goal is that \workname~can be a benchmark tool to evaluate the performance of different LVLMs.

\subsection{Broader Impact of \workname} 
We hope that \workname~can make a meaningful impact across several fields. First, \workname~can serve as a benchmark to support conventional HAR algorithms, including, but not limited to, evaluating multimodal algorithms, cross-subject approaches, and cross-domain methods. Additionally, it provides a benchmark for assessing the capabilities of current LLMs in action understanding and reasoning. Second, \workname~offers synchronous multimodal sensor data, making it easier for researchers and practitioners to explore and work with various sensors and tools. This feature makes it a valuable educational resource, serving as a standard dataset for teaching essential topics such as sensor fusion, data annotation, and multimodal reasoning.

\end{document}